\documentclass[9.5pt,article,compsoc]{IEEEtran}
\usepackage{import}
\usepackage{mystyle}
\usepackage{subfiles}

\usepackage{ragged2e}
\graphicspath{./figures}
\def\resultspath{figures/results-low-res}

\begin{document}
	\title{Learning Regional Attraction  for\\ Line Segment Detection}
	\author{Nan~Xue, 
		Song~Bai,
		Fu-Dong~Wang,
		Gui-Song~Xia, \\
		Tianfu~Wu,
		Liangpei~Zhang,
		Philip H.S. Torr
		\IEEEcompsocitemizethanks{
			\IEEEcompsocthanksitem 
			N. Xue, F.-D. Wang and L. Zhang are with State Key Lab. LIESMARS, Wuhan University, China.
			
			E-mail: \{xuenan, fudong-wang, zlp62\}@whu.edu.cn
			
			\IEEEcompsocthanksitem 	G.-S. Xia is with School of Computer Science and the State Key Lab. LIESMARS, Wuhan University, China. 
			
			E-mail: guisong.xia@whu.edu.cn
			
			\IEEEcompsocthanksitem S. Bai and P. Torr are with the University of Oxford, United Kingdom.
			
			E-mail: songbai.site@gmail.com, philip.torr@eng.ox.ac.uk
			\IEEEcompsocthanksitem T. Wu is with Dept. Electrical \& Computer Engineering, NC State University, USA.
			
			E-mail: tianfu\_wu@ncsu.edu
			
			Corresponding author: Gui-Song Xia (guisong.xia@whu.edu.cn)
		}
		
	}
	\IEEEtitleabstractindextext{
		\justify
		\begin{abstract}
			This paper presents \emph{regional attraction} of line segment maps, and hereby poses the problem of line segment detection (LSD) as a problem of region coloring.
			Given a line segment map, the proposed regional attraction first establishes the relationship between line segments and regions in the image lattice. Based on this, the line segment map is equivalently transformed to an attraction field map (AFM), which can be remapped to a set of line segments without loss of information. Accordingly, we develop an end-to-end framework to learn attraction field maps for raw input images, followed by a squeeze module to detect line segments. Apart from existing works, the proposed detector properly handles the local ambiguity and does not rely on the accurate identification of edge pixels.
			Comprehensive experiments on the Wireframe dataset and the YorkUrban dataset demonstrate the superiority of our method. In particular, we achieve an F-measure of $0.831$ on the Wireframe dataset, advancing the state-of-the-art performance by $10.3$ percent.		
		\end{abstract}
	}
	\maketitle
	
	\vspace{-3mm}
	\IEEEraisesectionheading{\section{Introduction}\label{sec:introduction}}
	
\begin{figure}
    \centering
    \subfigure[Input\label{fig:teaser-input}]{
    \includegraphics[width=0.35\linewidth]{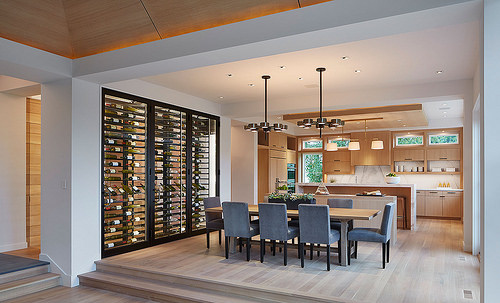}}
    \subfigure[Ours\label{fig:teaser-ours}]{
    \includegraphics[width=0.59\linewidth]{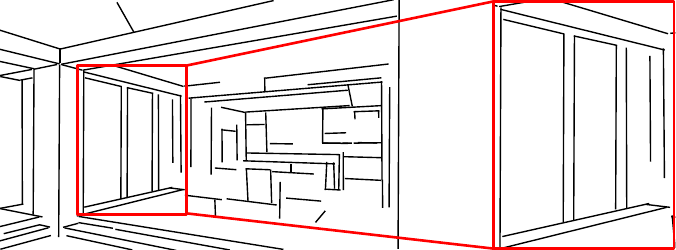}
    }\\
    \subfigure[Local Edge Map\label{fig:teaser-local-edge}]{
    \includegraphics[width=0.35\linewidth]{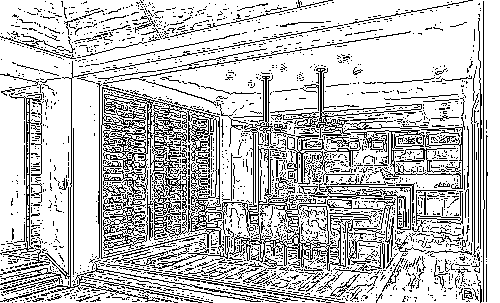}}
    \subfigure[MCMLSD~\cite{Almazan_2017_CVPR}\label{fig:teaser-mcmlsd}]{
    \includegraphics[width=0.59\linewidth]{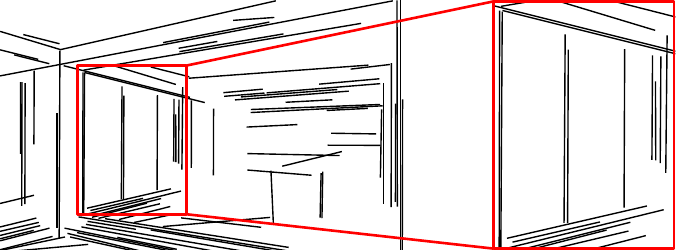}}\\
    \subfigure[Gradient Magnitude\label{fig:teaser-gradient}]{
    \includegraphics[width=0.35\linewidth]{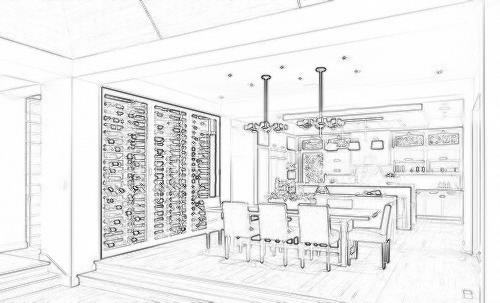}}
    \subfigure[LSD~\cite{VonGioi2010}\label{fig:teaser-lsd}]{\includegraphics[width=0.59\linewidth]{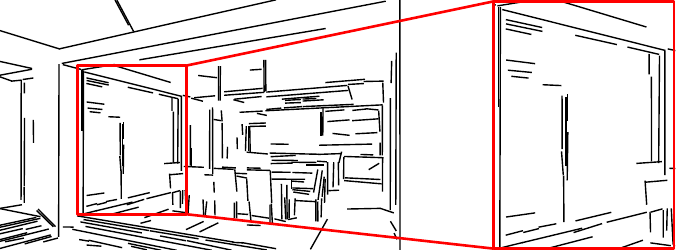}}\\
    \subfigure[Deep Edge Map\label{fig:teaser-deep-edge}]{
    \includegraphics[width=0.35\linewidth]{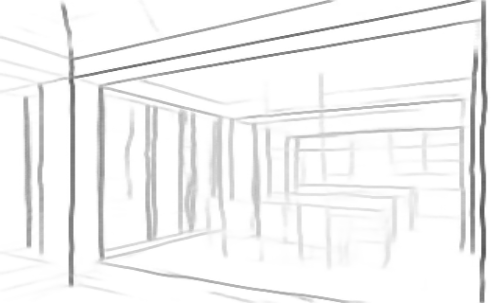}
    }
    \subfigure[DWP~\cite{Huang2018a}\label{fig:teaser-dwp}]{\includegraphics[width=0.59\linewidth]{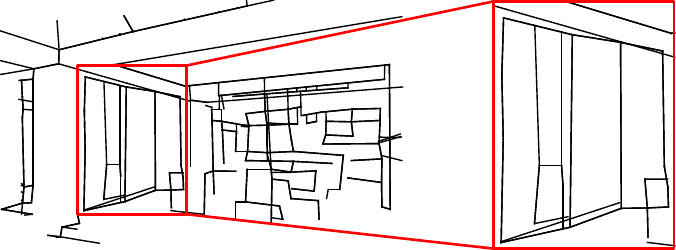}}
\vspace{-2mm}    
    \caption{Illustrative examples of different methods on an image (a). (b) shows our detected line segments. (c) and (d) present the locally estimated edge map and the result of MCMLSD~\cite{Almazan_2017_CVPR}. (e) and (f) present the gradient magnitude and the result of LSD~\cite{VonGioi2010}. (g) and (h) display the deep edge map and the result of Deep Wireframe Parser (DWP)~\cite{Huang2018a}. The rightmost column shows the close-up (in red) of detection results by different methods, which highlights the better accuracy of our proposed method. Best viewed in color.}
    \label{fig:my_label}
\end{figure}

	\IEEEPARstart{L}{ine} segment detection (LSD) is an important yet challenging low-level task in computer vision~\cite{BurnsHR86,VonGioi2010,Almazan_2017_CVPR}. LSD aims to extract visible line segments in scene images (see Figure~\ref{fig:teaser-input} and Figure~\ref{fig:teaser-ours}).
	The resulting line segments of an image provide a compact structural representation that facilitates many up-level vision tasks such as 3D reconstruction~\cite{Denis2008,FaugerasDMAR92}, image partitioning~\cite{Duan2015}, stereo matching~\cite{Yu2013}, scene parsing~\cite{Zoua,Zhao2013}, camera pose estimation~\cite{Xu2017}, and image stitching~\cite{XiangXBZ18}. 

	
	LSD is usually formulated as a heuristic search problem~\cite{VonGioi2010,Almazan_2017_CVPR} that groups or fits the edge pixels into several line segments. 
	The classical Hough transform (HT)~\cite{Ballard81}, as well as some HT-based variants~\cite{MatasGK00,YangGH11,ShiGRAC13,GuerreiroA12, Almazan_2017_CVPR}, takes locally estimated edge maps as input to fit straight lines in the first step and then estimates the endpoints of the line segments according to the density of the edge pixels on these straight lines. These methods suffer from the incorrect edge pixel identification (see Figure~\ref{fig:teaser-local-edge}) in the locally estimated edge maps, and often produce a number of false positive detections (see the result of MCMLSD~\cite{Almazan_2017_CVPR} in Figure~\ref{fig:teaser-mcmlsd}). 
	
	In contrast to HT-based approaches, Burn~\emph{et al.}~\cite{BurnsHR86} attempted to locally group edge cues into line segments. Following this, LSD~\cite{VonGioi2010} and Linelet~\cite{Cho2018} grouped pixels with high gradient magnitude values (\ie, edge pixels) into line segment proposals according to the gradient orientation. Once the line segment proposals were obtained, the validation processes based on the Helmholtz principle~\cite{DesolneuxMM00,DesolneuxMM03} were applied to reject false positive detections.
	However, edge pixels in low-contrast regions were prone to being omitted, thereby breaking a long line segment into several short ones. An example of LSD~\cite{VonGioi2010} and the corresponding gradient magnitude are given in Figure~\ref{fig:teaser-lsd} and Figure~\ref{fig:teaser-gradient} respectively.
	
	It is problematic for those methods to detect complete line segments while suppressing false alarms using traditional edge cues~\cite{Almazan_2017_CVPR,VonGioi2010,Cho2018}. Furthermore, the edge pixels can only approximately characterize the line segment as a set of connected pixels, also suffering from unknown multiscale discretization nuisance factors (\eg,~the classic zig-zag artifacts of line segments in digital images).
	
	In recent years, convolutional neural networks (ConvNets) have demonstrated a potential for going beyond the limitation of local approaches to detect edge pixels with global context. 
	The Holistically-nested Edge Detector (HED)~\cite{Xie2015a} used the fully convolutional network (FCN) architecture~\cite{LongSD15} for the first time to learn and detect edge maps for input images in an end-to-end manner. Later, many deep learning based edge detection systems were proposed~\cite{Maninis2016, Kokkinos2016,Liu2017a} and significantly outperformed traditional edge detectors~\cite{Kittler83,MarrH80,Canny86a, TorreP86}. 
	Benefiting from the advances in deep edge detection, the deep wireframe parser (DWP)~\cite{Huang2018a} transforms line segment detection into edge maps and junction detections with two ConvNets and then fuses detected junctions and edge pixels into line segments. As shown in Figure~\ref{fig:teaser-deep-edge}, the estimated edge maps can identify edge pixels better in regions with complicated appearances, thus pushing the performance bounds of LSD forward by a large margin. However, the over-smoothing effect of deep edge detection will lead to local ambiguity for accurate line segment detection. 
	In Figure~\ref{fig:teaser-dwp}, some detected line segments are misaligned because of the blurred edge responses.
	
	\begin{figure*}[htb!]
		\centering
		\includegraphics[width=.95\linewidth]{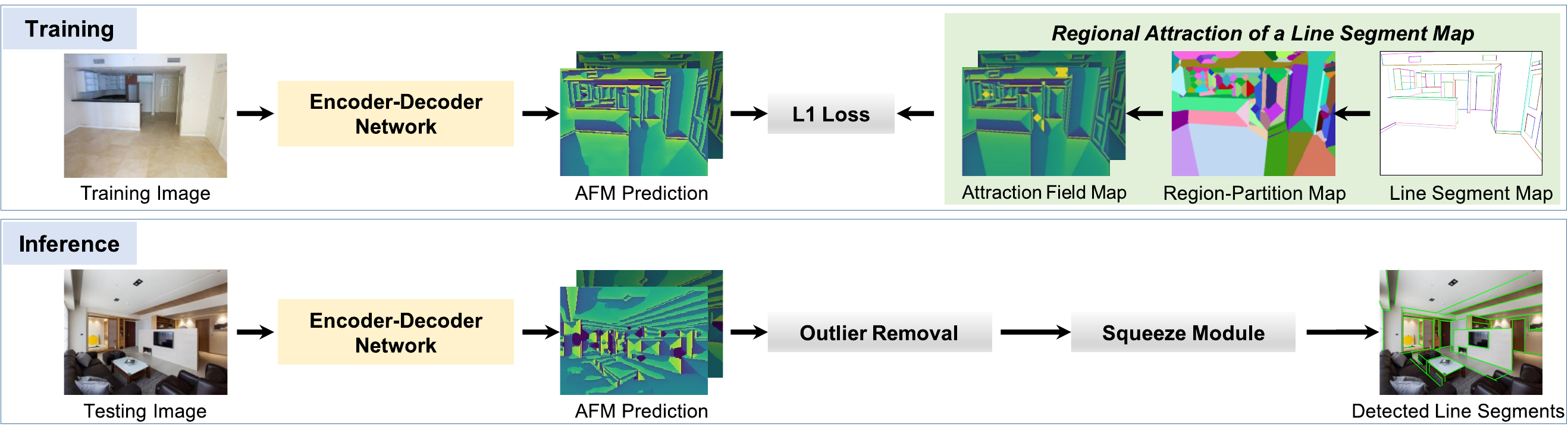}
		\vspace{-3mm}
		\caption{An illustration of the proposed regional attraction and line segment detection system. In the training phase, the annotated line segments of an image are equivalently represented by an attraction field map (AFM). Then, the image and corresponding AFM are feed into the encoder-decoder network for learning. In the inference phase, a testing image is passed into the trained network to obtain the AFM prediction. After removing the outliers and squeezing the predictions, the system outputs a set of line segments.}
		\label{fig:regional-representation}
	\end{figure*}
	
	In summary, most previous work~\cite{VonGioi2010,Almazan_2017_CVPR,Cho2018, Huang2018a} is built upon edge pixel identification and suffers from two main drawbacks: such work lacks elegant solutions to solve the issues caused by inaccurate or incorrect edge detection results (\eg, local ambiguity, high false positive detection rates and incomplete line segments) and requires carefully designed heuristics or extra contextual information to infer line segments from identified edge pixels.
	
	In this paper, we focus on a deep learning based LSD framework and propose a single-stage method that rigorously addresses the drawbacks of existing LSD approaches. Our method is motivated by the following observations:
	\begin{itemize}
		\item[-] The duality between regions and the contour (or the surface) of an object is well-known in computer vision~\cite{Martin2004a}.
		\item[-] All pixels in the image lattice should be involved in the formation of line segments in an image.
		\item[-] The recent remarkable progress led by deep learning based methods (\emph{e.g.},~U-Net~\cite{Ronneberger2015} and DeepLab V3+~\cite{Hou2018}) in semantic segmentation.
	\end{itemize} 
	
	Thus, the intuitive idea of this paper is that when bridging a line segment map and its spatial proximate regions, we can pose the problem of LSD as the problem of region coloring, and thus open the door to leveraging the best practices developed in state-of-the-art deep ConvNet based semantic segmentation methods to improve the performance of LSD.
	
	\subsection{Method Overview}
	Following this idea, we exploit the spatial relationship between pixels in the image lattice and line segments, and propose a new formulation termed \emph{regional attraction} for line segment detection (as shown in Figure~\ref{fig:regional-representation}).
	Our proposed regional attraction establishes the relation between 1D line segments and 2D regions of image lattices, and an induced representation characterizes the geometry of line segments by using edge pixels and non-edge pixels together.
	Compared with the previous formulation of line segment detection, our proposed \emph{regional attraction} can directly encode the geometric information of line segments without using the edge maps.
	
	By learning the regional attraction, our proposed line segment detector eliminates the limitations of edge pixel identification. 
	As shown in Figure~\ref{fig:my_label}, our method yields a much better result than several representative line segment detectors, especially in the gray region that has high-frequency textures.
	
	We establish the relationship between pixels and line segments by seeking the most ``attractive" line segment for every pixel in the image lattice. Suppose that there are $n$ line segments on an image lattice $\Lambda$, where the most attractive line segment for every pixel $\bfs{p}\in\Lambda$ is defined as the nearest line segment of pixel $\bfs{p}$. By applying this criterion, the pixels in the image lattice $\Lambda$ are partitioned into $n$ regions $\{R_i\}_{i=1}^n$, which form a region-partition map.
	Consequently, non-edge pixels are also involved to depict the geometry of  line segments.
	In detail, we use the shortest vector from every pixel $\bfs{p} \in R_i$ to its most ``attractive" line segment to characterize the geometric property of the line segment. As an example, if the pixel $\bfs{p} \in R_i$ can reach a point inside the line segment, the vector will simultaneously depict the location and normal direction of the line segment. Otherwise, the vector indicates the endpoint of the line segment. We term such vectors as the attraction vectors. The attraction vectors of every pixel $\bfs{p}$ together form an attraction field map (AFM).
	
	The format of attraction field maps is actually a two-dimensional feature map, which is compatible with convolutional neural networks. Therefore, regional attraction allows the problem of LSD to be transformed into a problem of region coloring. More importantly, thanks to  recent advances of deep learning based semantic segmentation methods, it is feasible to learn the attraction field map in an end-to-end manner. Once the attraction field map of an image can be estimated accurately, regional attraction is capable of recovering the line segment map in a nearly perfect manner via a simple and efficient squeeze module. The regional attraction can also be viewed as an intuitive expansion-and-contraction operation between 1D line segments and 2D regions: the region-partition map jointly expands all line segments into partitioned regions, and the squeeze module degenerates regions into line segments.
	
	Figure~\ref{fig:regional-representation} illustrates the pipeline of the proposed LSD framework based on an encoder-decoder neural network. 
	Specifically, we utilize a modified network based on DeepLab V3+~\cite{Hou2018} in our experiments to estimate the attraction field maps for line segment detection.
	In the training phase, the proposed regional attraction first forms a region-partition map and then generates ground truth of the attraction field map to supervise the training of the deep network. 
	In the testing phase, the attraction field map computed by the network is squeezed to output line segments. Compared with the preliminary version of Attraction Field Map (AFM)~\cite{AFMCVPR}, we further propose an outlier removal module based on the statistical priors of the training dataset, which significantly improves the performance of LSD. Besides, we find that better optimizer (\eg, the  Adam optimizer~\cite{Adam-Optimizer}) with adaptive learning rate decay can make ConvNets learn better attraction field maps. We name the enhanced version of the line segment detector as AFM++.  
	
	\subsection{Contributions}
	Our work makes the following  contributions to robust line segment detection, as
	\begin{itemize}
		\item A novel representation of line segments is proposed to bridge line segment maps and region-partition-based attraction field maps. To the best of our knowledge, it is the first work that utilizes this simple yet effective representation for LSD.
		\item With the proposed regional attraction, the problem of LSD is then solved by using a ConvNet without the necessity of identifying edge pixels.
		\item The proposed AFM++ obtains state-of-the-art performance on two widely used LSD benchmarks, including the Wireframe~\cite{Huang2018a} and YorkUrban~\cite{Denis2008} datasets. In particular, on the Wireframe dataset, AFM++ beats the current best-performing algorithm by 10.3 percent.
	\end{itemize}
	
	The reminder of this paper is organized as follows. Existing research related to our work is briefly reviewed in Section~\ref{sec:related_work}. In Section~\ref{sec:rep}, the details of the regional attraction for line segments are presented, followed by the definition of AFM++ in Section~\ref{sec:detection}. The experimental results and comparisons are given in Section~\ref{sec:experiments}. Finally, we conclude our paper in Section~\ref{sec:conclusion}.

	\section{Related Work}\label{sec:related_work}
	{
		\subsection{Benchmark Datasets for Line Segment Detection}
		Like many other vision problems, benchmark datasets are important for evaluating the performance of a line segment detector. However, the ill-posed definition of line segment detection brings some difficulties to create a perfect benchmark dataset for line segment detection. Specifically, the perception ambiguity will lead to some inconsistency for annotating line segments from images. The well-known BSDS dataset~\cite{Martin2004a} suffered from this issue in edge detection and they tried to use multi-source annotations to eliminate the ambiguity. For the problem of line segment detection, the existing benchmark datasets (\eg, the Wireframe dataset~\cite{Huang2018a} and the YorkUrban dataset~\cite{Denis2008}) tried to address this issue by using some priors of human perception or the scene geometry. Specifically, the line segments annotations of the Wireframe dataset~\cite{Huang2018a} are obtained by associating the salient scene structures. For the YorkUrban dataset~\cite{Huang2018a}, they use the vanishing points as a criterion to annotate line segments and each line segment is associated with one of the vanishing points. In this paper, we use the Wireframe dataset~\cite{Huang2018a} and the YorkUrban dataset~\cite{Denis2008} to evaluate our proposed line segment detector, and our proposed method consistently obtains the state-of-the-art performance on these two datasets that have different annotation rules.
	}
	\subsection{Detection based on Local Edge Cues}
	For a long time, hand-crafted low-level edge cues were extensively used in line segment detection. 
	The classical LSD baseline takes the output of an edge detector (\eg, Canny detector~\cite{Canny86a}) and then applies Hough transform~\cite{Ballard81} (HT) to fit infinite-long straight lines. Then, line segments are obtained by cutting these straight lines according to the density of the edge pixels on the lines. 
	Since the locally estimated edge maps suffer from a number of false positive edge pixels, it is challenging to detect line segments from input images robustly. 
	The incorrectly identified edge pixels will produce many spurious peaks in the Hough space, which will produce a number of false positive and false negative detections. 
	The progressive probabilistic Hough transform (PPHT)~\cite{MatasGK00} proposed a false detection control to improve the detection results of the classical Hough transform. Desolneux \emph{et al.}~\cite{DesolneuxMM00,DesolneuxMM03} addressed the issue of false detection by applying Helmholtz principle in line segment detection. In this method, the meaningful aligned line segments are retained as the final detections. Moreover, the distribution of peaks in the Hough space was studied in~\cite{FurukawaS03,XuSK15,XuSK15A,XuSK15B} to improve the performance of LSD. 
	Most recently, MCMLSD~\cite{Almazan_2017_CVPR} proposed to control the false detections by exploiting the distribution of edge pixels on the voted straight lines. 
	However, the performance of HT-based approaches still cannot achieve the satisfactory performance.
	
	In contrast to fitting line segments from edge pixels, Burn \emph{et al.}~\cite{BurnsHR86} found that the local gradient orientation is more robust to intensity variations than the gradient magnitude (and local edge maps). 
	Based on this, a perception grouping approach~\cite{BurnsHR86} was proposed to detect line segments without using Hough transform. Given a gray-scale image, adjacent pixels with similar gradient orientations are grouped to yield a set of line segments. Similar to the HT-based approaches, this approach also suffers from false positive detections. 
	Subsequently, a novel grouping approach based on Helmholtz principle~\cite{DesolneuxMM00,DesolneuxMM03} was proposed in~\cite{GioiJMR08}.
	Afterward, LSD~\cite{VonGioi2010} was proposed to improve the performance of line segment detection in both speed and accuracy. Benefiting from the development of Helmholtz principle in the problem of LSD, the grouping approaches can suppress false detections by applying an \emph{a-contrario} validation processes. 
	Nevertheless, it is still a challenge to detect complete line segments in low-contrast regions.
	To this end, the ASJ detector~\cite{Xue2017} was proposed to detect long line segments starting from detected junctions~\cite{Xia2014}. 
	However, that approach still suffers from the uncertainty caused by the image gradient. 
	Recently, Cho~\emph{et al.}~\cite{Cho2018} proposed a linelet-based  framework to address the problem of LSD. In this framework, pixels with large gradient magnitudes are grouped into linelets, and the line segment proposals are obtained by grouping the adjacent linelets. A probabilistic validation process is applied to reject false detections. To avoid incomplete results,  line segment proposals passed the validation are fed into an aggregation process to detect complete line segments.
	Similar to the HT-based approaches, the performance of perception grouping approaches also rely on whether the image gradient can reflect the edge information in a precise way.
	
	The performance of these line segment detectors depends on if the edge pixels can be correctly extracted. The edge maps (including image gradient maps) used for line segment detection are obtained from the local features, which are easily affected by the external imaging conditions ({\em e.g.},~noise and illumination). 
	Therefore, the local nature of these approaches poses a challenge to accurately extract line segments from images even with powerful validation processes. 
	Compared with the approaches based on local edge cues, our proposed method achieves robust line segment detection by learning more effective deep features. Moreover, our proposed detector only requires a simple criterion to reject false detections.
	
	\subsection{Deep Edge and Line Segment Detection}
	Recently, HED~\cite{Xie2015a} opened up a new era for edge perception in images by using ConvNets. The learned multi-scale and multi-level features effectively address the problem of false detection in the edge-like texture regions and approach human-level performance on the BSDS500 dataset~\cite{Martin2004a}. 
	From the perspective of binary classification, edge detection has been solved to some extent. 
	It then inspires researchers to upgrade the existing edge-based line segment detectors to deep-edge based line segment detectors.
	Convolutional Oriented Boundaries (COB)~\cite{Maninis2016,Maninis2018a} detector was proposed to get multi-scale oriented contours and region hierarchies from a single ConvNet. Since the oriented contours are adaptive to the input format (\ie, edge pixels and orientations) of fast LSD~\cite{VonGioi2010}, they can be used to address the issue of incomplete detection in LSD effectively.
	However, the edge maps estimated by ConvNets are usually over-smoothed, which leads to local ambiguities for accurate localization.
	In comparison to edge detection, deep learning based line segment detection has not yet been well investigated and  requires further exploration.
	
	Most recently, Huang~{\em et al.}~\cite{Huang2018a} took an important step toward this goal by collecting a large-scale dataset with high-quality line segment annotations and approaching the problem of line segment detection as two parallel tasks, \ie, {edge map detection} and {junction detection}. In the final step, the resulted edge map and junctions are merged to produce line segments. To the best of our knowledge, this is the first attempt to develop a deep learning based line segment detector. However, due to the sophisticated relation between edge map and junctions, inferring line segments from edge maps and junction cues in a precise way is still an open problem.
	
	Compared with this approach, our proposed formulation enables us to detect line segments from the attraction field maps instead of using edge maps and additional junction cues. The richer geometric information encoded in the attraction field maps facilitates line segment detection without considering the blurring effect of deep edge detectors.
	
	{
		Furthermore, learning signed distance functions has been widely and successfully used for representing 2D closed-boundaries~\cite{Andrew00a,EstellersZLOTB12,Yuan18} and 3D object surfaces~\cite{ZollhoferDIWSTN15,DeepSDF}.
		our proposed attraction field representation shares the similar spirit, but differs in two aspects: Our proposed method directly learns the attraction vectors instead of the distance maps, which can explicitly and accurately characterize the geometry of line segments, and thus eliminates the need of considering the approximation errors for numerical computation. And, our proposed formulation takes the pixels in the non-zero level sets (\ie, non-edge pixels) into accounts for achieving robust line segment detection.
	}
	\section{Regional Attraction}\label{sec:rep}
	In this section, we provide the details of regional attraction to characterize the line segments. 
	Concretely, we introduce a region-partition map to bridge the relationship between the line segments and regions in Section~\ref{sec:region-partition}. In Section~\ref{sec:attraction-field}, we utilize the attraction field map (AFM) to depict the 1D geometry by using all pixels in the image lattice. In Section~\ref{sec:squeeze}, we show that the attraction field map can be remapped into line segments by using a simple yet efficient squeeze module, which establishes the foundation of a deep learning based line segment detection system. Further analyses are given in Section~\ref{sec:RA-analysis}.
	
	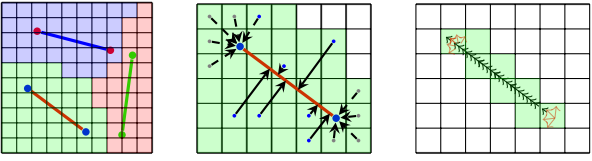
\begin{figure}[!tb]
		\centering
		\subfigure[Support regions\label{fig:pols-proposal}]{
			\begin{tikzpicture}[scale=0.2]
        \pgfmathsetmacro{\ymin}{0}
        \pgfmathsetmacro{\xmin}{0}
        \pgfmathsetmacro{\ymax}{10}
        \pgfmathsetmacro{\xmax}{10}	
        \node[circle, fill, inner sep=1.0pt, red] (A0) at (2.36,8.1) {};
        \node[circle, fill, inner sep=1.0pt, red] (A1) at (7.23,6.83) {};
        \node[circle, fill, inner sep=1.0pt, green] (B0) at (8,1.2) {};
        \node[circle, fill, inner sep=1.0pt, green] (B1) at (8.7,6.5) {};
        \node[circle, fill, inner sep=1.0pt, blue] (C0) at (1.73,4.29) {};
        \node[circle, fill, inner sep=1.0pt, blue] (C1) at (5.60,1.40) {};
        
        \draw [blue, very thick] (A0) -- (A1);	
        \draw [green, very thick] (B0) -- (B1);
        \draw [red, very thick] (C0) -- (C1);

        \foreach \i in {\xmin,\xmax} {
			\draw [thick, gray] (\i, \ymin) -- (\i, \ymax) node [below] at (\i, \ymin) {};
		}
		\foreach \i in {\ymin,\ymax} {
		\draw [thick, gray] (\xmin, \i) -- (\xmax, \i) node [below] at (\xmin, \i) {};
	    }        
        \draw [step=1.0, black] (\xmin, \ymin) grid (\xmax,\ymax);

        \fill [green, opacity=0.2] (0,0) -- (7,0) -- (7,3) -- (6,3) -- (6,5) -- (4,5) -- (4,6) -- (0,6) -- (0,0);
        \fill [red, opacity=0.2] (6,5) -- (6,3) -- (7,3) -- (7,0) -- (10,0) -- (10,10) -- (9,10) -- (9,9) -- (8,9) -- (8,8) -- (8,7) -- (8,6) -- (7,6) -- (7,5) -- (6,5);
        \fill [blue, opacity=0.2] (0,6) -- (4,6) -- (4,5) -- (7,5) -- (7,6) -- (8,6) -- (8,9) -- (9,9) -- (9,10) -- (0,10) -- (0,6);
        \end{tikzpicture}
		}
		\subfigure[Attraction vectors \label{fig:details}]{
			\begin{tikzpicture}[scale=0.33]
\pgfmathsetmacro{\ymin}{0}
\pgfmathsetmacro{\xmin}{0}
\pgfmathsetmacro{\ymax}{6}
\pgfmathsetmacro{\xmax}{7}	
\foreach \i in {\xmin,...,\xmax} {
		\draw [thick, gray] (\i, \ymin) -- (\i, \ymax) node [below] at (\i, \ymin) {};
	}
\foreach \i in {\ymin,\ymax} {
	\draw [thick, gray] (\xmin, \i) -- (\xmax, \i) node [below] at (\xmin, \i) {};
    }
\draw [step=1.0, black] (\xmin, \ymin) grid (\xmax,\ymax);
\node[circle, fill, inner sep=1.0pt, blue] (A) at (1.73,4.29) {};
\node[circle, fill, inner sep=1.0pt, blue] (B) at (5.60,1.40) {};
\draw [red, very thick] (A) -- (B);
\fill [green, opacity=0.2] (0,0) -- (7,0) -- (7,3) -- (6,3) -- (6,5) -- (4,5) -- (4,6) -- (0,6) -- (0,0);

\node[circle, fill, inner sep=0.5pt, gray](ea1) at (0.5,5.5) {};
\node[circle, fill, inner sep=0.5pt, gray](ea2) at (1.5,5.5) {};
\node[circle, fill, inner sep=0.5pt, gray](ea3) at (0.5,4.5) {};
\node[circle, fill, inner sep=0.5pt, gray](ea4) at (1.5,4.5) {};
\node[circle, fill, inner sep=0.5pt, gray](ea5) at (0.5,3.5) {};
\node[circle, fill, inner sep=0.5pt, blue](ea6) at (2.5,5.5) {};
\node[circle, fill, inner sep=0.5pt, gray](eb1) at (6.5,2.5) {};
\node[circle, fill, inner sep=0.5pt, gray](eb2) at (6.5,1.5) {};
\node[circle, fill, inner sep=0.5pt, gray](eb3) at (6.5,0.5) {};
\node[circle, fill, inner sep=0.5pt, gray](eb4) at (5.5,0.5) {};
\node[circle, fill, inner sep=0.5pt, blue](c3) at (1.5,1.5) {};
\node[circle, fill, inner sep=0.5pt, blue](d2) at (2.5,1.5) {};
\node[circle, fill, inner sep=0.5pt, blue](e4) at (3.5,3.5) {};
\node[circle, fill, inner sep=0.5pt, blue](f1) at (4.5,0.5) {};
\node[circle, fill, inner sep=0.5pt, blue](f2) at (4.5,1.5) {};	
\node[circle, fill, inner sep=0.5pt, blue](g3) at (5.5,4.5)  {};	
\draw [->,>=stealth, dashed, thick] (ea1) -- (A);
\draw [->,>=stealth, dashed, thick] (ea2) -- (A);
\draw [->,>=stealth, dashed, thick] (ea3) -- (A);
\draw [->,>=stealth, dashed, thick] (ea4) -- (A);
\draw [->,>=stealth, dashed, thick] (ea5) -- (A);
\draw [->,>=stealth, dashed, thick] (ea6) -- (A);
\draw [->,>=stealth, dashed, thick] (eb1) -- (B);
\draw [->,>=stealth, dashed, thick] (eb2) -- (B);
\draw [->,>=stealth, dashed, thick] (eb3) -- (B);
\draw [->,>=stealth, dashed, thick] (eb4) -- (B);
\draw [->,>=stealth, thick] (c3) -- ($(A)!(c3)!(B)$);
\draw [->,>=stealth, thick] (d2) -- ($(A)!(d2)!(B)$);
\draw [->,>=stealth, thick] (e4) -- ($(A)!(e4)!(B)$);
\draw [->,>=stealth, thick] (f1) -- ($(A)!(f1)!(B)$);
\draw [->,>=stealth, thick] (f2) -- ($(A)!(f2)!(B)$);
\draw [->,>=stealth, thick] (g3) -- ($(A)!(g3)!(B)$);
\end{tikzpicture}
		}
		\subfigure[Squeeze module\label{fig:squeeze}]{
			\begin{tikzpicture}[scale=0.33]
\pgfmathsetmacro{\ymin}{0}
\pgfmathsetmacro{\xmin}{0}
\pgfmathsetmacro{\ymax}{6}
\pgfmathsetmacro{\xmax}{7}	
\foreach \i in {\xmin,...,\xmax} {
		\draw [thick, gray, opacity=0] (\i, \ymin) -- (\i, \ymax) node [below] at (\i, \ymin) {};
	}
\foreach \i in {\ymin,\ymax} {
	\draw [thick, gray] (\xmin, \i) -- (\xmax, \i) node [below] at (\xmin, \i) {};
    }
\draw [step=1.0, black] (\xmin, \ymin) grid (\xmax,\ymax);
\node[inner sep=1.0pt, blue] (A) at (1.73,4.29) {};
\node[inner sep=1.0pt, blue] (B) at (5.60,1.40) {};
\coordinate (dir) at (-0.26708016,  0.19944745);
\node[inner sep=0.5pt] (a0) at (0.5, 0.5) {};
\node[inner sep=0.5pt] (a0p) at ($(A)!(a0)!(B)$){};
\draw[->] (a0p) -- ++(dir);
\node[inner sep=0.5pt] (a1) at (1.5, 0.5){};
\node[inner sep=0.5pt] (a1p) at ($(A)!(a1)!(B)$){};
\draw[->] (a1p) -- ++(dir);
\node[inner sep=0.5pt] (a2) at (2.5, 0.5){};
\node[inner sep=0.5pt] (a2p) at ($(A)!(a2)!(B)$){};
\draw[->] (a2p) -- ++(dir);
\node[inner sep=0.5pt] (a3) at (3.5, 0.5){};
\node[inner sep=0.5pt] (a3p) at ($(A)!(a3)!(B)$){};
\draw[->] (a3p) -- ++(dir);
\node[inner sep=0.5pt] (a4) at (4.5, 0.5){};
\node[inner sep=0.5pt] (a4p) at ($(A)!(a4)!(B)$){};
\draw[->] (a4p) -- ++(dir);
\node[inner sep=0.5pt] (a5) at (0.5, 1.5){};
\node[inner sep=0.5pt] (a5p) at ($(A)!(a5)!(B)$){};
\draw[->] (a5p) -- ++(dir);
\node[inner sep=0.5pt] (a6) at (1.5, 1.5){};
\node[inner sep=0.5pt] (a6p) at ($(A)!(a6)!(B)$){};
\draw[->] (a6p) -- ++(dir);
\node[inner sep=0.5pt] (a7) at (2.5, 1.5){};
\node[inner sep=0.5pt] (a7p) at ($(A)!(a7)!(B)$){};
\draw[->] (a7p) -- ++(dir);
\node[inner sep=0.5pt] (a8) at (3.5, 1.5){};
\node[inner sep=0.5pt] (a8p) at ($(A)!(a8)!(B)$){};
\draw[->] (a8p) -- ++(dir);
\node[inner sep=0.5pt] (a9) at (4.5, 1.5){};
\node[inner sep=0.5pt] (a9p) at ($(A)!(a9)!(B)$){};
\draw[->] (a9p) -- ++(dir);
\node[inner sep=0.5pt] (a10) at (0.5, 2.5){};
\node[inner sep=0.5pt] (a10p) at ($(A)!(a10)!(B)$){};
\draw[->] (a10p) -- ++(dir);
\node[inner sep=0.5pt] (a11) at (1.5, 2.5){};
\node[inner sep=0.5pt] (a11p) at ($(A)!(a11)!(B)$){};
\draw[->] (a11p) -- ++(dir);
\node[inner sep=0.5pt] (a12) at (2.5, 2.5){};
\node[inner sep=0.5pt] (a12p) at ($(A)!(a12)!(B)$){};
\draw[->] (a12p) -- ++(dir);
\node[inner sep=0.5pt] (a13) at (3.5, 2.5){};
\node[inner sep=0.5pt] (a13p) at ($(A)!(a13)!(B)$){};
\draw[->] (a13p) -- ++(dir);
\node[inner sep=0.5pt] (a14) at (4.5, 2.5){};
\node[inner sep=0.5pt] (a14p) at ($(A)!(a14)!(B)$){};
\draw[->] (a14p) -- ++(dir);
\node[inner sep=0.5pt] (a15) at (5.5, 2.5){};
\node[inner sep=0.5pt] (a15p) at ($(A)!(a15)!(B)$){};
\draw[->] (a15p) -- ++(dir);
\node[inner sep=0.5pt] (a16) at (1.5, 3.5){};
\node[inner sep=0.5pt] (a16p) at ($(A)!(a16)!(B)$){};
\draw[->] (a16p) -- ++(dir);
\node[inner sep=0.5pt] (a17) at (2.5, 3.5){};
\node[inner sep=0.5pt] (a17p) at ($(A)!(a17)!(B)$){};
\draw[->] (a17p) -- ++(dir);
\node[inner sep=0.5pt] (a18) at (3.5, 3.5){};
\node[inner sep=0.5pt] (a18p) at ($(A)!(a18)!(B)$){};
\draw[->] (a18p) -- ++(dir);
\node[inner sep=0.5pt] (a19) at (4.5, 3.5){};
\node[inner sep=0.5pt] (a19p) at ($(A)!(a19)!(B)$){};
\draw[->] (a19p) -- ++(dir);
\node[inner sep=0.5pt] (a20) at (5.5, 3.5){};
\node[inner sep=0.5pt] (a20p) at ($(A)!(a20)!(B)$){};
\draw[->] (a20p) -- ++(dir);
\node[inner sep=0.5pt] (a21) at (1.5, 4.5){};
\node[inner sep=0.5pt] (a21p) at ($(A)!(a21)!(B)$){};
\draw[->] (a21p) -- ++(dir);
\node[inner sep=0.5pt] (a22) at (2.5, 4.5){};
\node[inner sep=0.5pt] (a22p) at ($(A)!(a22)!(B)$){};
\draw[->] (a22p) -- ++(dir);
\node[inner sep=0.5pt] (a23) at (3.5, 4.5){};
\node[inner sep=0.5pt] (a23p) at ($(A)!(a23)!(B)$){};
\draw[->] (a23p) -- ++(dir);
\node[inner sep=0.5pt] (a24) at (4.5, 4.5){};
\node[inner sep=0.5pt] (a24p) at ($(A)!(a24)!(B)$){};
\draw[->] (a24p) -- ++(dir);
\node[inner sep=0.5pt] (a25) at (5.5, 4.5){};
\node[inner sep=0.5pt] (a25p) at ($(A)!(a25)!(B)$){};
\draw[->] (a25p) -- ++(dir);
\draw [red,opacity=0.5,->] (A) -- ++(-0.35643939549437625, 0.3506436329660124);
\draw [red,opacity=0.5,->] (A) -- ++(-0.09336950361691489, 0.4912047798976827);
\draw [red,opacity=0.5,->] (A) -- ++(-0.4928682127552036, 0.08414823144601036);
\draw [red,opacity=0.5,->] (A) -- ++(-0.36924274696406295, 0.33713468201066615);
\draw [red,opacity=0.5,->] (A) -- ++(-0.42070033194990575, -0.2702059042605086);
\draw [red,opacity=0.5,->] (A) -- ++(0.26843774609657967, 0.42183074386605374);
\draw [red,opacity=0.5,->] (5.6, 1.4) -- ++(-0.3166188951286314, -0.38697864960166045);
\draw [red,opacity=0.5,->] (5.6, 1.4) -- ++(-0.4969418673368095, -0.0552157630374233);
\draw [red,opacity=0.5,->] (5.6, 1.4) -- ++(-0.35355339059327384, 0.3535533905932737);
\draw [red,opacity=0.5,->] (5.6, 1.4) -- ++(0.055215763037423087, 0.4969418673368095);
\fill[green,opacity=0.2] (1.0, 4.0) rectangle (2.0, 5.0);
\fill[green,opacity=0.2] (2.0, 4.0) rectangle (3.0, 5.0);
\fill[green,opacity=0.2] (2.0, 3.0) rectangle (3.0, 4.0);
\fill[green,opacity=0.2] (3.0, 3.0) rectangle (4.0, 4.0);
\fill[green,opacity=0.2] (3.0, 2.0) rectangle (4.0, 3.0);
\fill[green,opacity=0.2] (4.0, 2.0) rectangle (5.0, 3.0);
\fill[green,opacity=0.2] (4.0, 1.0) rectangle (5.0, 2.0);
\fill[green,opacity=0.2] (5.0, 1.0) rectangle (6.0, 2.0);
\end{tikzpicture}
		}
		\vspace{-3mm}
		\caption{A toy example illustrating a line segment map with 3 line segments, including (a) the region-partition map with 3 regions, (b) selected attraction vectors and (c) the squeeze module for obtaining line segments.}
		\label{fig:lsmap}
		\vspace{-2mm}
	\end{figure}
	
	\subsection{Region-Partition Map}\label{sec:region-partition}
	Let $\Lambda$ be an image lattice ({\em e.g.}, $800\times 600$). A line segment is denoted by $\bfs{l}_i=(\mathbf{x}^s_i, \mathbf{x}^e_i)$ with the two endpoints being $\mathbf{x}^s_i$ and $\mathbf{x}^e_i$  (non-negative real-valued positions as sub-pixel precision is used in annotating line segments) respectively. The set of line segments in a 2D image lattice is denoted by $L=\{\bfs{l}_1, \cdots, \bfs{l}_n\}$. For simplicity, we term the set $L$ as a line segment map.  
	Figure~\ref{fig:lsmap} illustrates a line segment map with 3 line segments in a $10\times 10$ image lattice. 
	
	The region-partition map assigns each pixel $\bfs{p}\in\Lambda$ to the nearest line segment in $L$.
	To this end, we use a point-to-line-segment distance function. Considering a pixel $\bfs{p}\in \Lambda$ and a line segment $\bfs{l}_i=(\mathbf{x}_i^s, \mathbf{x}_i^e)\in L$, we first project the pixel $\bfs{p}$ onto the straight line passing through $\bfs{l}_i$ in the continuous geometry space. If the projection point is not on the line segment, we use the closest endpoint of the line segment as the projection point. Then, we compute the Euclidean distance between the pixel and the projection point. Formally, we define the distance between $\bfs{p}$ and $\bfs{l}_i$ by
	\begin{equation}
	\begin{split}
	d(\bfs{p}, \bfs{l}_i) &= 
	\min_{t\in[0,1]} d(\bfs{p}, \bfs{l}_i; t) \\
	& = \min_{t\in [0,1]}||\mathbf{x}_i^s + t\cdot (\mathbf{x}_i^e - \mathbf{x}_i^s) - \bfs{p}||_2^2, \\
	t_p^* &= \arg\min_{t\in[0,1]} d(\bfs{p}, \bfs{l}_i;t),
	\end{split}
	\end{equation}
	where the projection point is the original point-to-line projection point if $t^*_p\in(0, 1)$, and is the closest endpoint if $t^*_p=0$ or $1$.  
	
	Then, the region in the image lattice for the line segment $\bfs{l}_i$ is defined by
	\begin{equation}
	R_i =\{\bfs{p}\, |\, \bfs{p}\in \Lambda; d(\bfs{p}, \bfs{l}_i) < d(\bfs{p}, \bfs{l}_j), \forall j\neq i, \bfs{l}_j\in L\}.
	\end{equation}
	It is straightforward to see that $R_i\cap R_j=\emptyset$ and $\cup_{i=1}^n R_i=\Lambda$, {\em i.e.}, all $R_i$'s form a partition of the image lattice. Figure~\ref{fig:pols-proposal} illustrates the region partition of line segments in a toy example. Denote by $R=\{R_1, \cdots, R_n\}$ the region-partition map for a line segment map $L$. 
	
	\subsection{Attraction Field Map for Line Segments}\label{sec:attraction-field}
	The region-partition map defines a region for each line segment. Consider the region $R_i$ associated with the line segment $\bfs{l}_i$. For each pixel $\bfs{p}\in R_i$, its projection point $\bfs{p}'$ on $\bfs{l}_i$ is defined by
	\begin{equation}
	\bfs{p}' = \mathbf{x}_i^s + t_p^*\cdot (\mathbf{x}_i^e - \mathbf{x}_i^s).
	\end{equation}
	Then, we define the 2D attraction (or the projection vector) of the pixel $\bfs{p}$ in the support region $R_i$ as
	\begin{equation} \label{eq:def-attraction-vector}
	\mathbf{a}_i(\bfs{p}) = \bfs{p}' - \bfs{p},
	\end{equation}
	where the attraction vector is perpendicular to the line segment if $t^*_p\in(0, 1)$ (see Figure~\ref{fig:details}). The attraction mapping function in Equation~\eqref{eq:def-attraction-vector} is applied over the image lattice as
	\begin{equation} \label{eq:def-attraction-vector-on-image}
	\begin{split}
	\mathbf{a}: & \Lambda \to \mathbb{R}^2\\
	& \bfs{p} \mapsto \mathbf{a}_i(\bfs{p}), ~ ~\text{if} ~~ \bfs{p} \in R_i.
	\end{split}
	\end{equation}
	We term the mapping defined in Equation~\eqref{eq:def-attraction-vector-on-image} as the attraction field of the line segment map $L$. For simplicity, we denote the attraction field map (AFM) of $L$ as $A=\{\mathbf{a}(p)\, |~p\in \Lambda\}$ by enumerating all the pixels in $\Lambda$.
	
	Figure~\ref{fig:regional-representation} shows examples of the $x$- and $y$-component of an attraction field map. It should be mentioned here that the attraction field map can be regarded as a variant of distance transform~\cite{distance-transform}. Generally, the distance transform is applied to binary images, where a pixel inside foreground regions is changed to measure its minimal distance to the boundary. Specially in our scenario, we use AFM to explicitly encode the geometric relationship between pixels and line segments.
	
	Compared with the edge map (or image gradient map) used in previous work (\emph{e.g.},~DWP~\cite{Huang2018a}, LSD~\cite{VonGioi2010} and Linelet~\cite{Cho2018}), the advantages of attraction field map can be summarized as follows:
	\begin{itemize}
		\item The edge map only approximately characterizes the line segments with very few pixels, which results in zig-zag effects. In contrast, our proposed AFM depicts the geometry of line segments in a more precise way with redundantly sampling over the line segments.
		\item Because each line segment is associated with a well-defined support region, 
		our proposed representation does not need to consider the blurring effects for the nearly distributed parallel line segments. 
	\end{itemize}
	
	Next, we will show how to remap the attraction field map into a set of line segments.
	
	\subsection{Squeeze Module}\label{sec:squeeze}
	The squeeze module groups the attraction vectors that are adjacent to a set and the non-perpendicular vectors are used as a condition of terminating the grouping process for resulting line segments. 
	Given an attraction field map $A$, we can compute the real-valued projection point for each pixel $\mathbf{p}$ in the lattice as
	\begin{equation}
	\mathbf{v}(\mathbf{p}) = \mathbf{p}+\mathbf{a}(\mathbf{p}), 
	\end{equation}
	and its corresponding discretized point in the image lattice as
	\begin{equation}
	\mathbf{v}_{\Lambda}(\bfs{p}) = \lfloor \mathbf{v(p)} + 0.5 \rfloor,
	\end{equation}
	where $\lfloor \cdot \rfloor$ represents the floor operation, and $v_{\Lambda}(\bfs{p})\in \Lambda$. 
	In addition, the attraction field map provides the normal direction (if the projected point $\mathbf{v}(\bfs{p})$ is inside) of the line segment going through the point $\bfs{v}(\bfs{p})$ by
	\begin{equation}
	\phi(\mathbf{p}) = \text{arctan2}(\mathbf{a}_y(\bfs{p}),\mathbf{a}_x(\bfs{p})),
	\end{equation}
	where $\mathbf{a}_x{(\bfs{p})}$ and $\mathbf{a}_y{(\bfs{p})}$ are the x- and y- components of the vector $\bfs{a}(\bfs{p})$ respectively.
	
	Then, the attraction vectors are rearranged according to their discretized projecting points, which results in a sparse map for recording the locations of possible line segments. For notation simplicity, such a sparse map is termed a line proposal map in which each pixel $\bfs{q}\in \Lambda$ collects the attraction field vectors whose discretized projection points are $\bfs{q}$. The candidate set of attraction field vectors collected by a pixel $\bfs{q}$ is then defined by 
	\begin{equation}
	\mathcal{C}(\bfs{q}) = \{\bfs{a}(\bfs{p}) \, | \, \bfs{p}\in \Lambda, \bfs{v}_{\Lambda}(\bfs{p})=\bfs{q} \},
	\end{equation}
	where $\mathcal{C}(\bfs{q})$'s are usually non-empty for a sparse set of pixels $\bfs{q}$'s which correspond to points on the line segments. An example of the line proposal map is shown in Figure~\ref{fig:squeeze}, which projects the pixels of the support region for a line segment into pixels near the line segment.
	
	With the line proposal map, our squeeze module utilizes an iterative and greedy grouping strategy to fit line segments in the spirit of the region growing algorithm used in~\cite{VonGioi2010}. The pseudocode of the squeeze module is given in Algorithm \ref{alg:squeeze}.
	\begin{itemize}
		\item Given the current set of active pixels each of which having a non-empty candidate set of attraction field vectors, we randomly select a pixel $q$ and one of its attraction field vector $\bfs{a}(\bfs{p})\in \mathcal{C}(\bfs{q})$. The tangent direction of the selected attraction field vectors $\bfs{a}(\bfs{p})$ is used as the initial direction of the line segment passing the pixel $\bfs{q}$. 
		\item Then, we search the local observation window centered at $\bfs{q}$ ({\em e.g.,} a $3\times 3$ window is used in this paper) to find the attraction field vectors that are aligned with $\bfs{a}(\bfs{p})$ with an angular distance less than a threshold $\tau$ ({\em e.g.,} $\tau=10^\circ$ used in this paper). 
		\begin{itemize}
			\item If the search fails, we discard $\bfs{a}(\bfs{p})$ from $\mathcal{C}(\bfs{q})$, and further discard the pixel $\bfs{q}$ if $\mathcal{C}(\bfs{q})$ becomes empty.
			\item Otherwise, we grow $\bfs{q}$ into a set and update its direction by averaging the aligned attraction  vectors. The aligned attraction  vectors will be marked as used (and thus made inactive for the next round of search). For the two endpoints of the set, we recursively apply the greedy search algorithm to grow the line segment.
		\end{itemize}
		\item Once terminated, we obtain a candidate line segment $\bfs{l}_q=(\mathbf{x}_q^s, \mathbf{x}_q^e)$ with the support set of real-valued projection points. We fit the minimum outer rectangle using the support set. We verify the candidate line segment by checking the aspect ratio between the width and length of the approximated rectangle with respect to a predefined threshold to ensure the approximated rectangle is ``thin enough". If the checking fails, we mark the pixel $\bfs{q}$ inactive and release the support set to be active again.   
	\end{itemize}
	
	\begin{algorithm}
		\renewcommand{\algorithmicrequire}{\textbf{Input:}}
		\renewcommand{\algorithmicensure}{\textbf{Output:}}
		\newcommand*\Let[2]{\State #1 $\gets$ #2}
		
		\caption{Squeeze Module}
		\label{alg:squeeze}
		\begin{algorithmic}[1] 
			\Require{The attraction field map $A$}
			
			\State Generating the line proposal map 
			$$
			\mathcal{Q} = \{\bfs{q}  | \mathcal{C}(\bfs{q}) \neq \emptyset,~ \forall \bfs{q} \in \Lambda \}.
			$$
			
			\State
			Initialize the status $\mathcal{S}(\bfs{p})$ for every pixel $\bfs{p}\in \Lambda$ by,
			$$\mathcal{S}(\bfs{p}) \gets \left\{
			\begin{array}{cc}
			0 & \mathtt{if~~v_{\Lambda}(\bfs{p}) \notin \Lambda} \\
			1 & \mathtt{otherwise}
			\end{array}
			\right. .
			$$
			
			\Let{$L$}{$\emptyset$}
			\For{$\bfs{p} \in \Lambda$ with $\mathcal{S}(\bfs{p})=1$}
			\Let{$\theta_0$}{$(\phi(\bfs{p}) + \frac{\pi}{2}) \mod \pi $}
			\Procedure{R $\gets$ region\_grow}{$\mathbf{q}$}
			\If{$\mathcal{S}(p') = 0~~ \forall \mathbf{a}(\bfs{p}') \in \mathcal{C}(\bfs{q})$}
			\State	Exit
			\EndIf
			\Let{$R$}{$\{\mathbf{v}(\bfs{p})\}$}
			\State{Initialize $\theta$, $R$ and $\mathcal{S}$ from $\mathbf{a}(\bfs{q}')\in \mathcal{C}(\bfs{p})$ and $\theta_0$}
			\If{Initialization failed}
			Return $\emptyset$
			\EndIf
			\For{$\bfs{q}' \in R_{\Lambda}$} \Comment{{\scriptsize $R_\Lambda$ is the set of discretized points in $R$}}
			\For{$\mathbf{a}(\bfs{p}') \in \mathcal{C}(\mathbf{q}'')~~ \forall \mathbf{q}'' \in \mathcal{N}(\mathbf{q}')$}
			\Let{$\theta'$}{$(\phi(\mathbf{p'})+\frac{\pi}{2}) \mod \pi $ }
			\Let{$R'$}{$\emptyset$}
			\If{$\text{dis}(\theta, \theta'))<\tau$}
			\State{Average $\theta$ with $\theta'$}
			\Let{$\mathcal{S}(\mathbf{q'})$}{0}
			\Let{$R'$}{$R' \cup \{\mathbf{v(p')}\}$}
			\EndIf
			\EndFor
			\EndFor
			\Let{$R$}{$R\cup R'$}
			\EndProcedure
			\State Fitting a rectangle $(\mathbf{x}_1,\mathbf{x}_2, w)$ from the point set $R$
			\If{$r = {w/\left\|\mathbf{x}_1-\mathbf{x}_2\right\|} < \epsilon$}
			\Let{$ L $}{$L \cup \{\mathbf{l}_i = (\mathbf{x}_1,\mathbf{x}_2)\}$} 
			\State{Return $L$}
			\Else
			\Let{$\mathcal{S}(\mathbf{p'})$}{1},~~ $\forall \mathbf{v(p')} \in R$
			\State{Return $L$}
			\EndIf
			\EndFor
			\Ensure{A set of line segments $L = \left\{(\mathbf{x}_i^s, \mathbf{x}_i^e)\right\}_{i=1}^N$ }
		\end{algorithmic}
	\end{algorithm}
	
	\subsection{Verifying Duality and Scale Invariance}\label{sec:RA-analysis}
	So far, we have established a dual representation to depict the geometry of line segment maps in the image lattice. Given a line segment map $L$ defined over the image lattice $\Lambda$, we are able to compute the corresponding attraction field map and then squeeze the AFM back to a set of line segments. In this section, we are going to verify the duality between line segments and the corresponding attraction field map. Furthermore, the scale invariance of regional attraction representations is verified.
	
\begin{figure}[!tb]
		\centering
		\includegraphics[width=\linewidth]{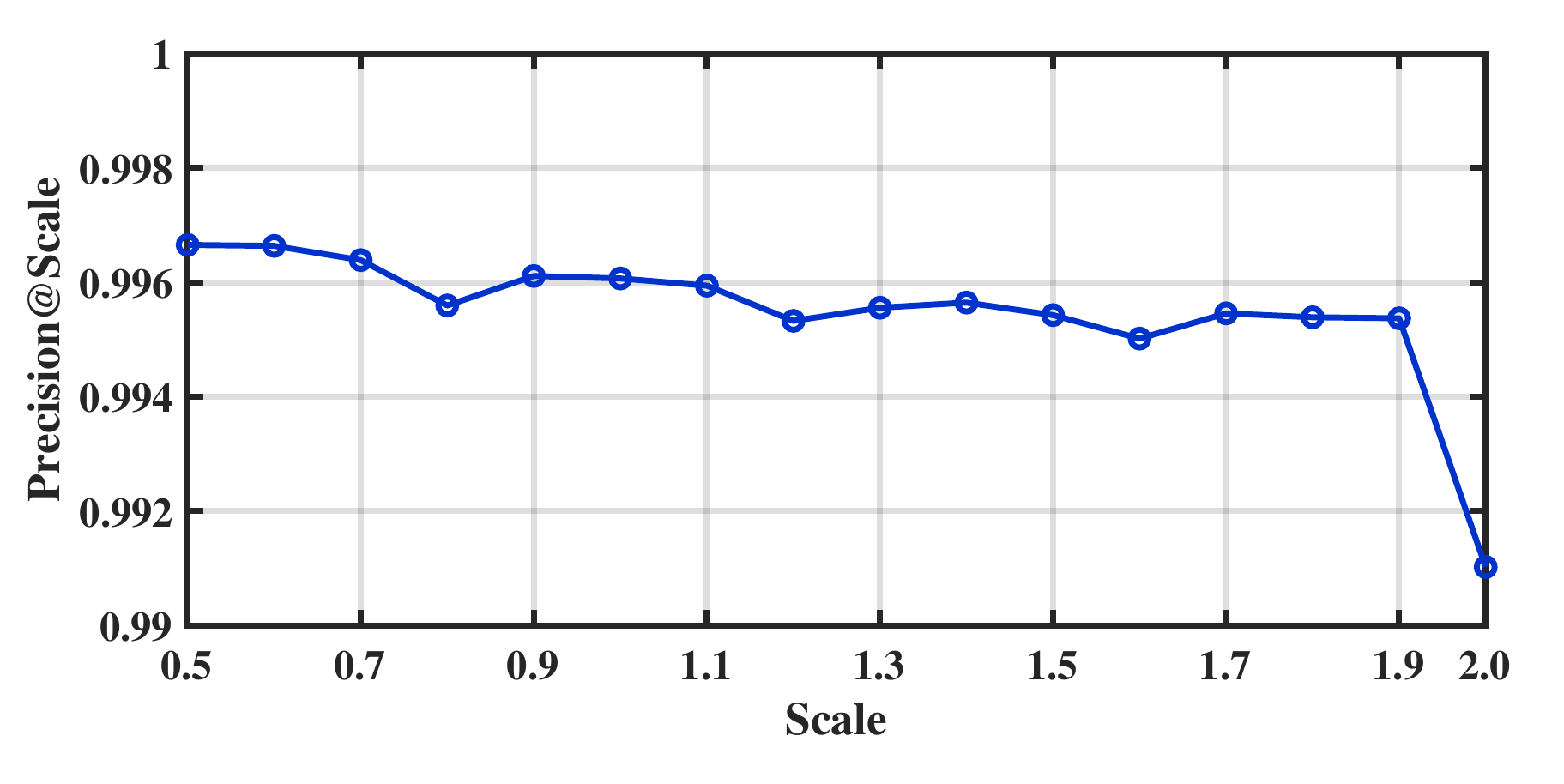}
		\includegraphics[width=\linewidth]{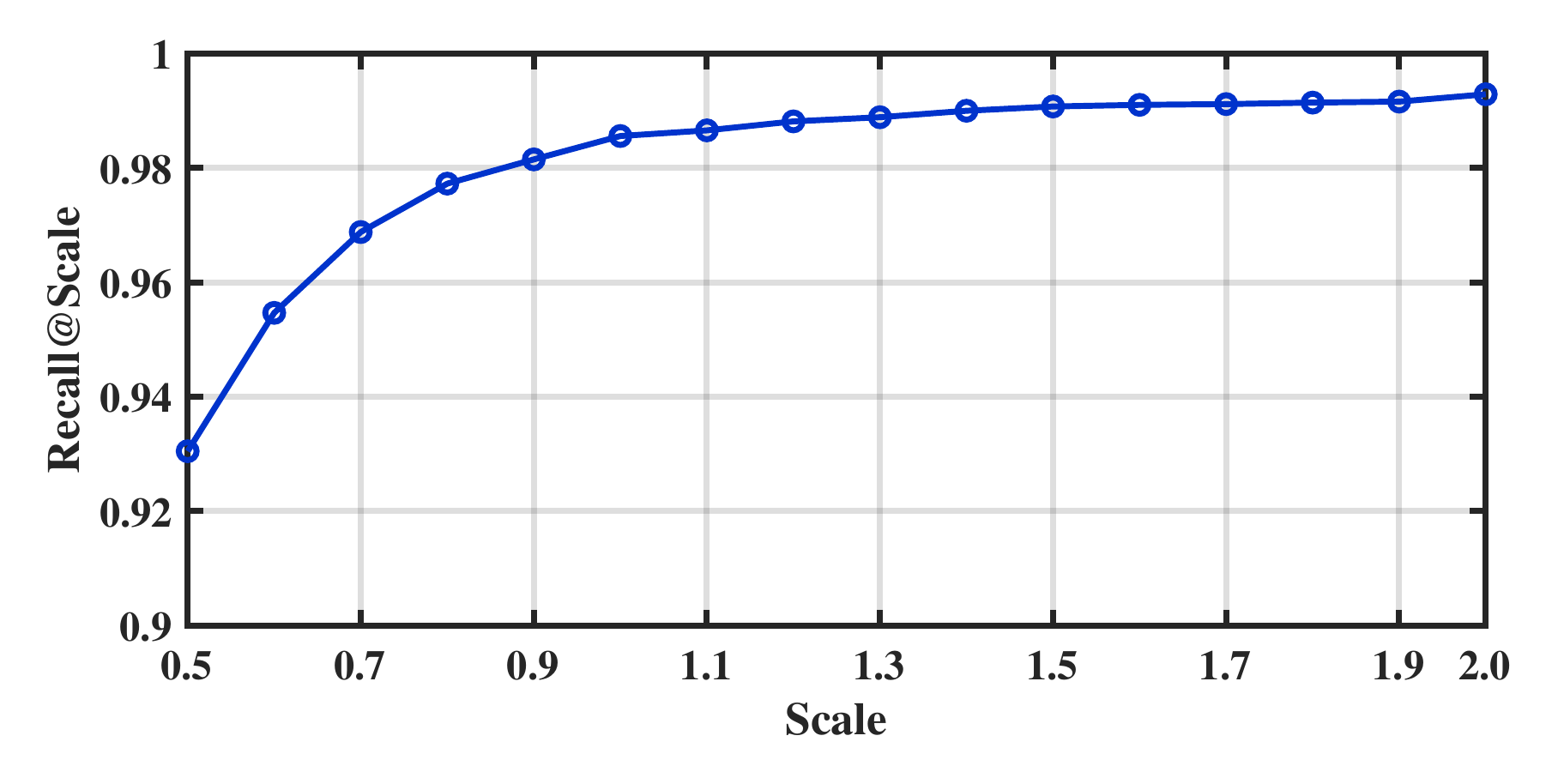}\\
		\vspace{-4mm}
		\caption{Verification of the duality between line segment maps and attraction field maps, and its scale invariance.}
		\label{fig:scaled-pr}
		\vspace{-2mm}
	\end{figure}

	We test the proposed regional attraction on the training split of the Wireframe dataset~\cite{Huang2018a}. We first compute the attraction field map for each annotated line segment map and then compute the estimated line segment map by using the squeeze module. 
	The verification is executed across multiple scales, varying from $0.5$ to $2.0$ with a step size of $0.1$. 
	The scale factor is used to control the size of attraction field maps.
	The estimated line segment maps are evaluated by measuring the precision and recall following the protocol provided along with the dataset. Figure~\ref{fig:scaled-pr} shows the precision-recall curves. The average precision and recall rates are above $0.99$ and $0.93$ respectively, thus verifying the duality between line segment maps and corresponding region-partition based attractive field maps, as well as the scale invariance of the duality. It is noteworthy that the precision will drop with the scale increases, which is probably caused by the fixed window size of $3\times 3$. When the scale increases, it is possible to induce more noisy attraction vectors, which will increase the probability to produce a bit more false positives. Despite this, the precision can be kept high as long as the attraction vectors are accurate.
	
	Therefore, the problem of LSD can be posed as a problem of region coloring without sacrificing the performance too much (the gap is negligible).
	With the formulation of regional attraction, our goal is to learn ConvNets to infer the attraction field maps for input images, which we will expand in the next section.
	
	\section{Deep Line Segment Detector} \label{sec:detection}
	In this section, we present the details of learning ConvNets for line segment detection. The proposed system takes the image $I$ as input and outputs $M$ line segments $L=\{\bfs{l}_j\}_{j=1}^{M}$.
	
	\subsection{AFM Parameterization}
	Denote by $D_{raw}=\{ (I_i, L_i)\}_{i=1}^N$  the provided training dataset consisting of $N$ pairs of raw images and annotated line segment maps. We first compute the AFM for each training image, then obtain the dual training dataset $D=\{ (I_i, \mathbf{a}_i); i=1,\cdots, N\}$. 
	
	\paragraph*{Numerical Stability and Scale Invariant Normalization}
	To make the AFMs insensitive to the sizes of raw images, we adopt a simple normalization scheme. For an AFM  $\bfs{a}$ with the height $H$ and the width $W$, the size-normalization is done by
	\begin{equation}\label{eq:scale-normalize}
	\bfs{a}_x := \bfs{a}_x/W,~~\bfs{a}_y := \bfs{a}_y/H,
	\end{equation}
	where $\bfs{a}_x$ and $\bfs{a}_y$ are the components of $\bfs{a}$ along $x$ and $y$ axes respectively. However, the size-normalization will make the values in $\bfs{a}$ quite small, which leads to numerically unstable training. We apply a point-wise invertible value stretching transformation for the size-normalized AFM as
	\begin{equation}\label{eq:log-normalize}
	z' := S(z) = -{\rm sign}(z)\cdot\log(|z|+\varepsilon),
	\end{equation}
	where $\varepsilon$ is set to $1\mathrm{e}{-6}$ to avoid $\log(0)$. The inverse function $S^{-1}(\cdot)$ is defined as
	\begin{equation}
	\label{eq:inverse-mapping}
	z := S^{-1}(z') = {\rm sign}(z') e^{(-|z'|)}.
	\end{equation} 
	For notation simplicity, denote by $\mathcal{R}(\cdot)$ the composite reverse function comprised of Equation~\eqref{eq:scale-normalize} and Equation~\eqref{eq:log-normalize}. We still denote by $D=\{ (I_i, \mathbf{a}_i); i=1,\cdots, N\}$ the final training dataset. 
	
	\subsection{Inference} 
	Denote by $f_{\Theta}(\cdot)$ a ConvNet with the parameters $\Theta$. As illustrated in Figure~\ref{fig:regional-representation}, for an input image $I_{\Lambda}$, the inference process of the proposed system is defined by 
	\begin{align}
	\hat{\mathbf{a}} & = f_{\Theta}(I_{\Lambda}) \\
	\hat{L} & = Squeeze(Inlier(\mathcal{R}(\hat{\mathbf{a}}))),     \label{eq:new-inference}
	\end{align}
	where $\hat{\mathbf{a}}$ is the predicted attraction field map for the input image (the size-normalized and value-stretched one).
	The $Inlier(\cdot)$ operator is designed to filter out inaccurate attraction vectors.
	$Squeeze(\cdot)$ denotes the squeeze module and $\hat{L}$ is the inferred line segment map. 
	
\begin{figure}[!tb]
		\centering
		\includegraphics[width=0.97\linewidth]{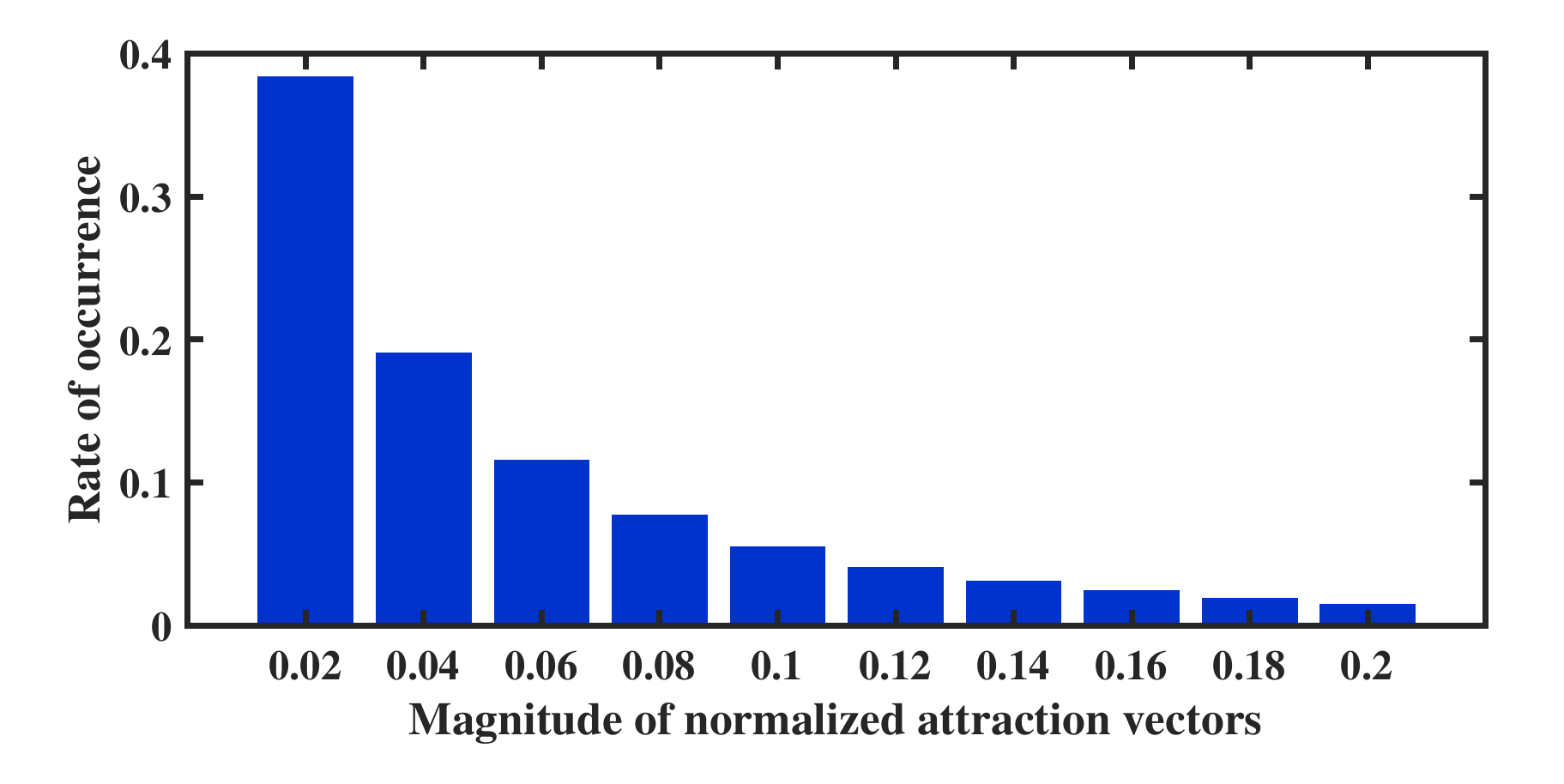}
		\vspace{-4mm}
		\caption{Distribution of magnitudes for the size-normalized attraction vectors in the training split of the Wireframe dataset.}
		\label{fig:distribution-afm}
	\end{figure}

	\paragraph*{Distribution of Regional Attraction and Outlier Removal} 
	Since not all the pixel predictions are accurate enough in practice, it is reasonable to remove potential outliers and only feed inliers to the squeeze module for better line segment detection.
	Meanwhile, our proposed regional attraction can depict every line segment with a relatively large region, the line segments can be precisely characterized even if we throw away some of attraction vectors.
	For the sake of computational efficiency, we analyze the magnitude of size-normalized attraction vectors on the training split of the Wireframe dataset~\cite{Huang2018a} in Figure~\ref{fig:distribution-afm}. 
	The magnitude of most attraction vectors are small than $0.02 \times \min\{H, W\}$. 
	Moreover, the networks should learn the vectors with small magnitude more accurately since a large penalty will be implicitly induced by using Equation~\eqref{eq:log-normalize}. 
	
	Observing this fact, we can filter out the outliers by using the magnitude of vectors without incurring any extra computational cost. Specifically, the $Inlier(\cdot)$ operator in Equation~\eqref{eq:new-inference} only retains the attraction vectors by
	\begin{equation}
	Inlier(R(\hat{\bfs{a}})) = \left\{\tilde{\bfs{a}} ~~| \tilde{\bfs{a}}\in \mathcal{R}(\hat{\bfs{a}}) \left\|\tilde{\bfs{a}}\right\|\leq \gamma\right\},
	\end{equation}
	where $\gamma$ is set to $0.02\times\min\{H,W\}$ according to the above discussion.
	
	\subsection{An \emph{a-trous} Residual U-Net} 
	Benefiting from our novel formulation, the problem of LSD can be addressed with the state-of-the-art encoder-decoder networks that are widely used in dense prediction tasks. However, the existing encoder-decoder architectures are usually designed to predict a down-sampled dense map due to the characteristics of tasks. For the problem of LSD, we expect to learn high-resolution attraction field maps to preserve the geometric information as much as possible. We achieve this by changing the stride of the conv1 layer to $1$ in U-Net architecture to ensure that the output feature map has the same size as the input image. Based on this, we  further adopt the ResBlock~\cite{He2016} and ASPP~\cite{Chena} modules to improve the learning ability of U-Net, which is termed as \emph{a-trous} Residual U-Net. 
	
	\begin{table}
		\centering
		\caption{Network architectures we use for attraction field learning. $\{\}$ and $[]$ represent the double conv in U-Net and the residual block, respectively. Inside the brackets are the shape of convolution kernels. The suffix $*$ represents the bi-linear up-sampling operator with a scaling factor of $2$. The number outside the brackets is the number of stacked blocks on a stage.}
		\vspace{-3mm}
		\label{tab:net-arch}
		\scriptsize
		\begin{tabular}{c|c|c}
			\hline
			stage  & U-Net & \emph{a-trous} Residual U-Net\\\hline
			c1 &  \doubleconv{3}{64} & $3\times3, 64$, stride 1  \\\hline
			\multirow{4}{*}{c2} &   $2\times 2$ max pool, stride 2 & $3\times 3$ max pool, stride 2 \\\cline{2-3}
			&  \doubleconv{3}{128} & \resconv{64}{64}{256}{3} \\\hline
			\multirow{3}{*}{c3} &   $2\times 2$ max pool, stride 2 &  \multirow{2}{*}{\resconv{128}{128}{512}{4}} \\\cline{2-2}
			&  \doubleconv{3}{256} &  \\\hline
			\multirow{3}{*}{c4} &   $2\times 2$ max pool, stride 2 & \multirow{2}{*}{\resconv{256}{256}{1024}{6}} \\\cline{2-2}
			&  \doubleconv{3}{512} &  \\\hline
			\multirow{3}{*}{c5} &   $2\times 2$ max pool, stride 2 & \multirow{2}{*}{ \resconv{512}{512}{2048}{3}} \\\cline{2-2}
			&  \doubleconv{3}{512} & \\\hline
			\multirow{4}{*}{d4} &  \multirow{4}{*}{\doubleconv[*]{3}{256}} & ASPP\\\cline{3-3}
			&   & \deconv{512}{512}{512}\\\hline
			d3 &  \doubleconv[*]{3}{128} & \deconv{256}{256}{256}\\\hline
			d2 & \doubleconv[*]{3}{64} & \deconv{128}{128}{128}\\\hline
			d1 & \doubleconv[*]{3}{64} & \deconv{64}{64}{64} \\\hline
			output & \multicolumn{2}{c}{$1\times 1$, stride 1, w.o. BN and ReLU}
			\\\hline
		\end{tabular}
	\end{table}
	Table~\ref{tab:net-arch} shows the configurations of U-Net and \emph{a-trous} Residual U-Net. The network consists of $5$ encoder  and $4$ decoder stages indexed by $c1, \ldots, c5$ and $d1, \ldots, d4$ respectively.
	
	\begin{itemize}
		\item For U-Net, the double conv operator, which contains two convolution layers, is applied and denoted as $\{\cdot\}$. The $\{\cdot\}*$ operator of $d_i$ stage upscales the output feature map of its last stage, then we concatenate it with the feature map of $c_i$ stage before applying the double conv operator. 
		\item For the \emph{a-trous} Residual U-Net, we replace the double conv operator with the Residual block, denoted as $[\cdot]$. In contrast to ResNet, we use the plain convolution layer with a $3\times 3$ kernel and a stride of $1$. Similar to $\{\cdot\}*$, the operator $[\cdot]*$ also takes the input from two sources and upscales the feature of the first input source. The first layer of $[\cdot]*$ contains two parallel convolution operators to reduce the depth of feature maps, then we concatenate them for the subsequent computations. In $d_4$ stage, we use 4 ASPP operators with the output channel size equal to $256$ and a dilation rate of $1,6,12,18$, then concatenate their outputs. The output stage is a $1\times 1$ convolution with a stride of $1$ without batch normalization~\cite{BN-IoffeS15} and ReLU~\cite{ReLU-NairH10} for the AFM prediction.
	\end{itemize}
	
	\subsection{Training and Testing} 
	We follow the standard deep learning protocol to estimate the parameters $\Theta$. We adopt the $l_1$ loss function in training, defined as
	\begin{equation}
	\ell(\hat{\bfs{a}}, \bfs{a}) = \sum_{(x,y)\in \Lambda} \|{\bf{a}(x,y)}-{\bf{\hat{a}}(x,y)}\|_1.
	\end{equation}
	
	\paragraph*{Baseline Implementation} 
	We train the networks from scratch on the training set of the Wireframe dataset~\cite{Huang2018a}. To make a fair comparison, we follow the standard data augmentation strategy from~\cite{Huang2018a} to enrich the training samples with image domain operations including mirroring and flipping upside-down. 
	The Adam optimizer is used here for training with the default settings in PyTorch ($\beta_1 = 0.9$ and $\beta_2 = 0.99$) and the initial learning rate is set to $0.001$. 
	We train all of the networks with $200$ epochs and the learning rate is decayed with the factor of $0.1$ after $180$ epochs. 
	In the training phase, we resize the images  to $320\times 320$ and then generate the attraction field maps from the resized line segment annotations to form the mini batches. As discussed in Section \ref{sec:rep}, the rescaling step with reasonable factors will not affect the results. The mini-batch sizes for the two networks are $16$ and $4$ respectively due to GPU memory limitations.
	
	In the inference stage, a test image is also resized to $320\times 320$ as the input of the network. Then, we use the squeeze module to convert the learned regional attraction into line segments. Since the line segments are insensitive to scale, we can directly resize them to original image size without sacrificing accuracy. The squeeze module is implemented with C++ on CPU.
	
	\section{Experiments} \label{sec:experiments}
	In this section, we evaluate the proposed line segment detector and compare with existing state-of-the-art line segment detectors~\cite{AFMCVPR,Huang2018a, Cho2018, Almazan_2017_CVPR, VonGioi2010} on the Wireframe dataset~\cite{Huang2018a} and YorkUrban dataset~\cite{Denis2008}. The source code of this paper will be released at \url{https://cherubicxn.github.io/afmplusplus/}.
	
	\subsection{Datasets and Evaluation Metrics}
	\paragraph*{Wireframe Dataset} The Wireframe dataset~\cite{Huang2018a} was proposed for line segment detection and junction detection. The images in this dataset are all taken in indoor scenes (\eg,~kitchens  and bedrooms) and outdoor man-made environments (\eg,~yards and houses). To the best of our knowledge, this dataset is the largest dataset (containing 5000 training samples and 462 testing samples)  with high-quality line segment annotations to date. The average resolution of the images in this dataset is $480\times 405$.
	Since this dataset focuses on scene structures, the line segments on the boundary of irregular or curved objects (\eg,~pillows and sofa) are not annotated. 
	In this paper, we train our line segment detector on the training split of this dataset and evaluate the performance on the testing split for comparison.
	
	\paragraph*{YorkUrban Dataset} The YorkUrban dataset~\cite{Denis2008} was initially proposed for edge-based Manhattan frame estimation and consists of $102$ images (45 indoor and 57 outdoor) with a size of $640\times 480$. The dataset is randomly split into a training set and a testing set with $51$ images each. For each image in this dataset, the ground truth line segments are annotated with sub-pixel precision. Since this dataset was designed for Manhattan world estimation, some of the line segments that are not associated with any vanishing point are not annotated. In this paper, we only use the testing split of this dataset for evaluation and performance comparison. We do not train or fine tune the model on this dataset. 
	
	\paragraph*{Evaluation Protocol}
	We follow the evaluation protocol from the DWP~\cite{Huang2018a} to make a comparison.
	First, the proposed method is evaluated on the testing split of  Wireframe dataset~\cite{Huang2018a}. To validate the ability of generalization, we also evaluate it on the YorkUrban dataset~\cite{Denis2008}.
	All the methods are evaluated quantitatively using precision and recall following~\cite{Huang2018a, Martin2004a}. The precision rate indicates the proportion of positive detections among all of the detected line segments while recall reflects the fraction of detected line segments among all in the scene. The detected and ground-truth line segments are digitized into the image domain and we define the ``positive detection'' pixel-wised. The line segment pixels within $0.01$ of the image diagonal are regarded as positive. After obtaining the precision (P) and recall (R), we compare the performance of algorithms using the F-measure
	$F = 2\cdot \frac{P\cdot R}{P+R}$.
	
	\subsection{Main Results for Comparison}
	We compare our proposed method with AFM~\cite{AFMCVPR}, DWP\footnote{\label{foot:wireframe}\url{https://github.com/huangkuns/wireframe}}~\cite{Huang2018a},
	Linelet\footnote{\label{foot:linelet}\url{https://github.com/NamgyuCho/Linelet-code-and-YorkUrban-LineSegment-DB}}~\cite{Cho2018}, 
	the Markov Chain Marginal Line Segment Detector\footnote{\label{foot:mcmlsd}\url{http://www.elderlab.yorku.ca/resources/}} (MCMLSD)~\cite{Almazan_2017_CVPR} and the Line Segment Detector (LSD)\footnote{\label{foot:lsd}\url{http://www.ipol.im/pub/art/2012/gjmr-lsd/}}~\cite{VonGioi2010}.
	The source codes of those methods are obtained from the links provided by the respective authors. 
	
	\paragraph*{Threshold Configuration}
	In our proposed method, we use the aspect ratio to filter out false detections. Here, we vary the threshold of the aspect ratio in the range $(0,1]$ with the step size $\Delta \tau = 0.02$. For comparison, the LSD~\cite{VonGioi2010} is evaluated with the $-\log(\text{NFA})$ in $0.01\times\{1.75^0, \ldots, 1.75^{19}\}$ for \emph{a-contrario} validation where $\text{NFA}$ is the number of false alarms. In addition, Linelet~\cite{Cho2018} uses the same thresholds as the LSD to filter out false detections. For MCMLSD~\cite{Almazan_2017_CVPR}, we use the top-$K$ detected line segments for evaluation. 
	With regard to the evaluation of DWP~\cite{Huang2018a}, we follow the default threshold setting for junction detection and line heat map binarization. In detail, the confidence threshold for both the junction localization and the junction orientation are set to $0.5$. The thresholds for line heat map binarization are set to $[2, 6, 10, 20, 30, 50, 80, 100, $ $150, 200, 250, 255]$ to detect line segments.
	
	\begin{table}[t]
		\centering
		\caption{Comparison of the F-measure with the state-of-the-art methods on the Wireframe and YorkUrban datasets. The last column reports the average inference speed (\emph{frames-per-second}, FPS) on the Wireframe dataset.}
		\vspace{-3mm}
		\small
		\label{tab:fmeasure-and-fps}
		\begin{tabular}{|l|c|c|c|}
			\hline
			Methods & \makecell{Wireframe\\ dataset} & \makecell{YorkUrban\\ dataset}  & FPS\\
			\hline
			\hline
			LSD~\cite{VonGioi2010} & 0.647 & 0.591 & \textbf{19.6} \\
			MCMLSD~\cite{Almazan_2017_CVPR} & 0.566 & 0.564 & 0.2 \\
			Linelet~\cite{Cho2018} & 0.644 & 0.585 & 0.14\\
			DWP~\cite{Huang2018a} & 0.728 & 0.627 & 2.24 \\
			\hline
			\hline
			AFM (U-Net)~\cite{AFMCVPR} & {0.753} & {0.639} & \textbf{10.3} \\
			AFM (\emph{a-trous})~\cite{AFMCVPR} & {0.774} & {0.647} & {6.6} \\
			AFM++ (\emph{a-trous}) & {\bf 0.823} & {\bf 0.672} & 8.0 \\
			\hline
		\end{tabular}
	\end{table}
	
	\begin{figure}[!tb]
		\centering
		\subfigure[PR curves on the Wireframe dataset\label{fig:pr-curve-wireframe}]{
			\includegraphics[width=.87\linewidth]{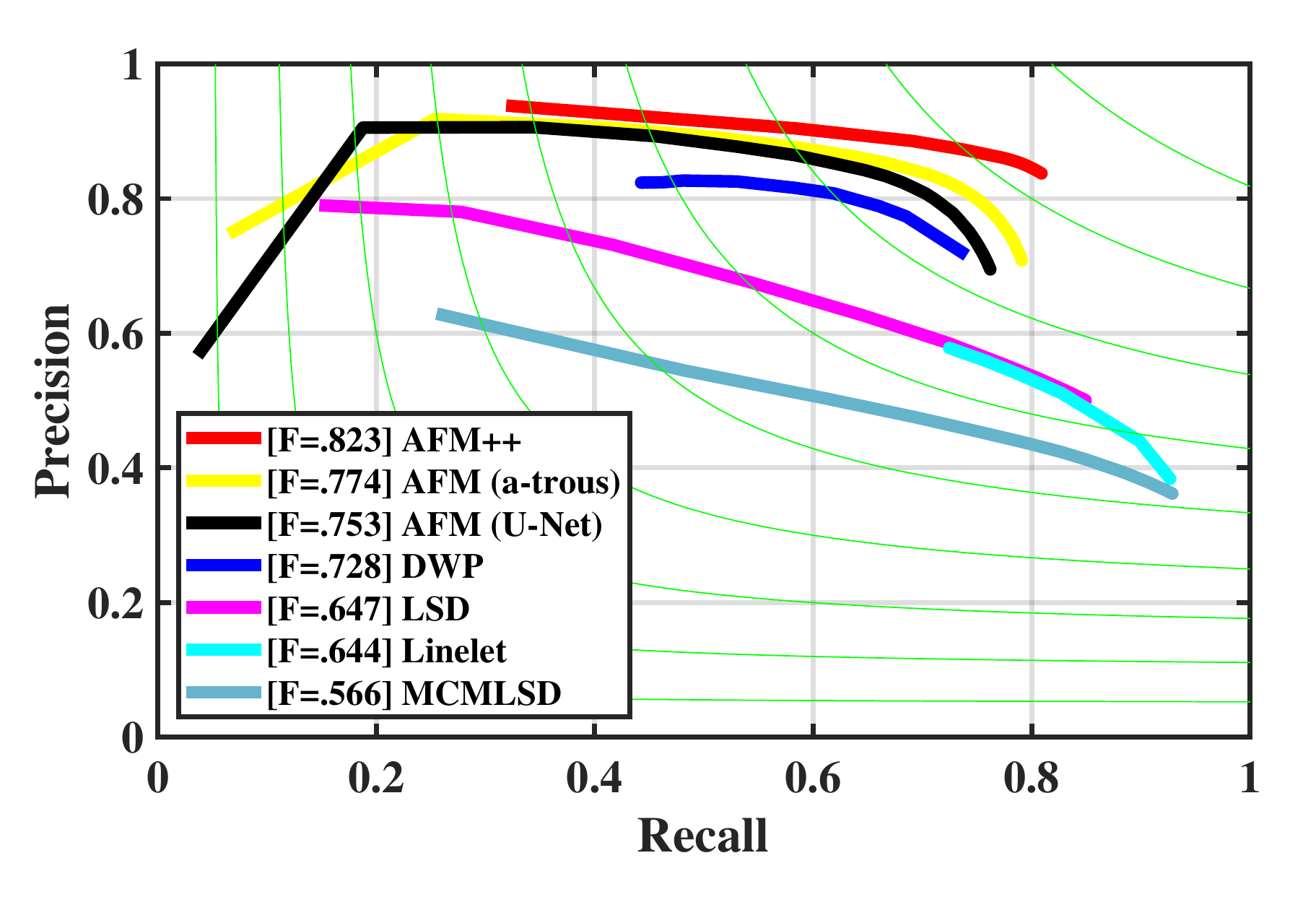}}
		\vspace{-3mm}
		\subfigure[PR curves on the YorkUrban dataset\label{fig:pr-curve-york}]{    \includegraphics[width=.87\linewidth]{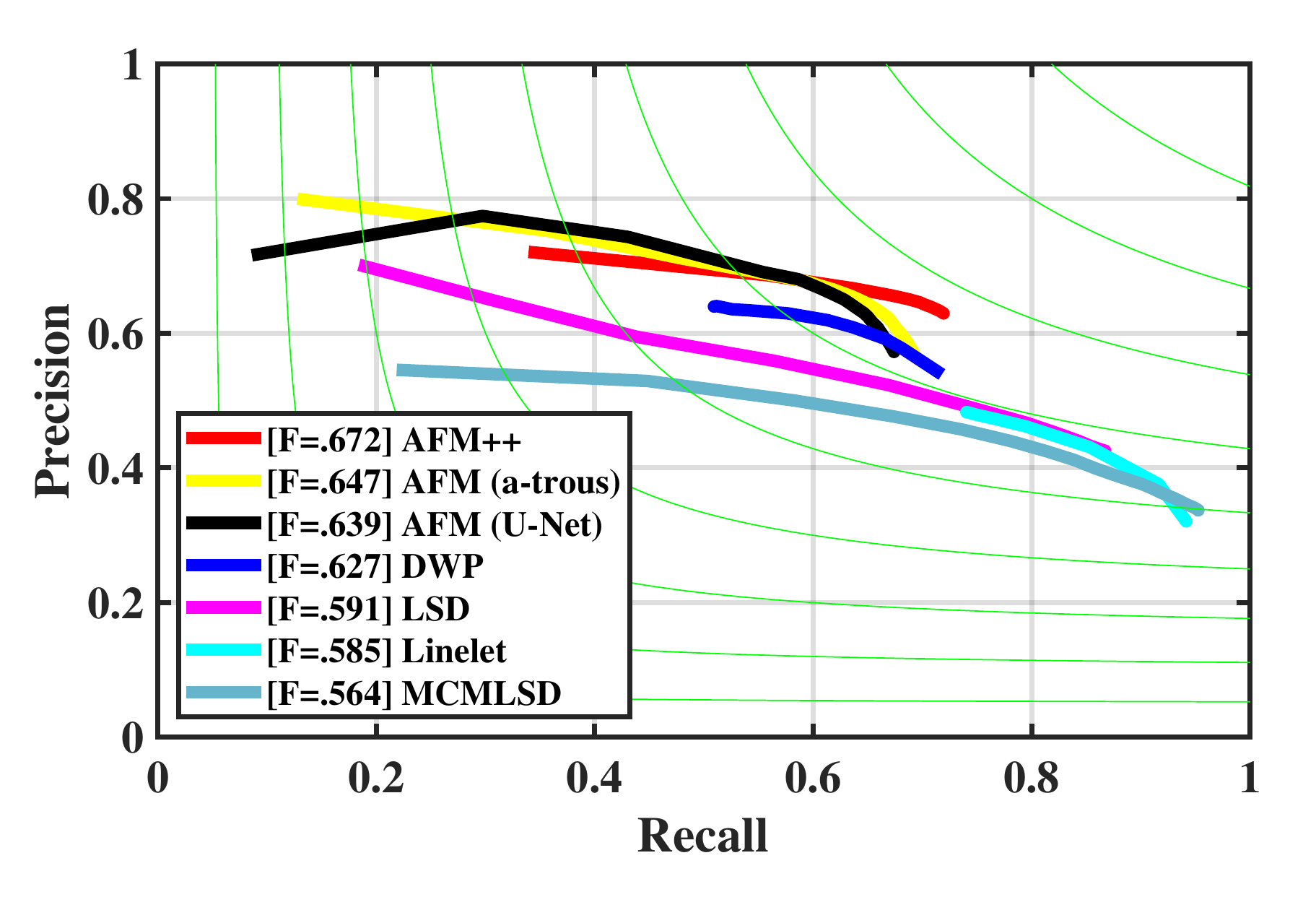}
		}
		\caption{The PR curves of different line segment detection methods on the Wireframe dataset~\cite{Huang2018a} and YorkUrban dataset~\cite{Denis2008}.}
		\vspace{-3mm}
	\end{figure}
	
	\begin{figure*}[ht!]
\centering
\def\outsize{0.13\textheight}
\def\xshift{-4mm}
\hspace{\xshift}
\begin{tikzpicture}
\node[rotate=90] (Ours) at (0,0) {LSD};
\draw[opacity=0] (-\outsize*0.01,-\outsize*0.5) rectangle (\outsize*0.01,\outsize*0.5);
\end{tikzpicture}
\includegraphics[height=\outsize]{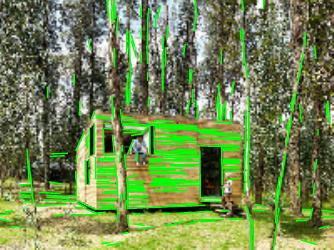}
\includegraphics[height=\outsize]{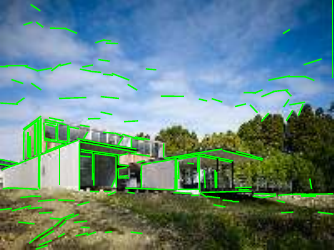}
\includegraphics[height=\outsize]{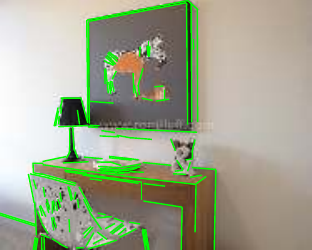}
\includegraphics[height=\outsize]{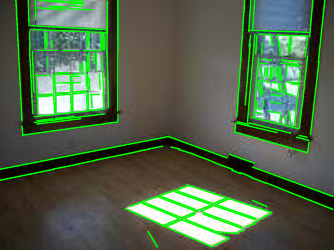}
\\
\hspace{\xshift}
\begin{tikzpicture}
\node[rotate=90] (Ours) at (0,0) {MCMLSD};
\draw[opacity=0] (-\outsize*0.01,-\outsize*0.5) rectangle (\outsize*0.01,\outsize*0.5);
\end{tikzpicture}
\includegraphics[height=\outsize]{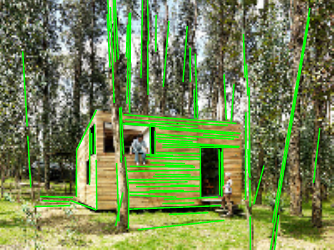}
\includegraphics[height=\outsize]{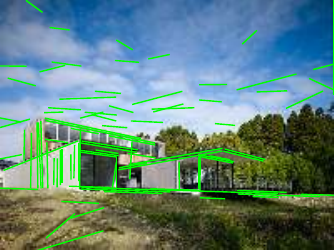}
\includegraphics[height=\outsize]{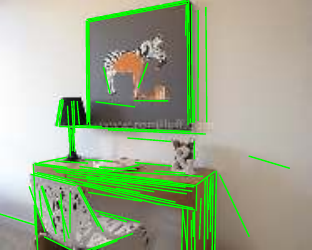}
\includegraphics[height=\outsize]{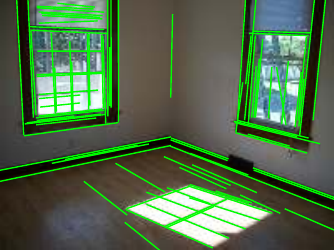}
\\
\hspace{\xshift}
\begin{tikzpicture}
\node[rotate=90] (Ours) at (0,0) {Linelet};
\draw[opacity=0] (-\outsize*0.01,-\outsize*0.5) rectangle (\outsize*0.01,\outsize*0.5);
\end{tikzpicture}
\includegraphics[height=\outsize]{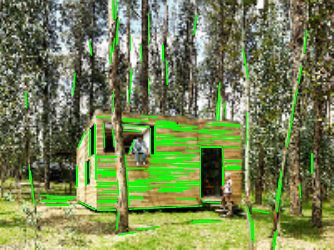}
\includegraphics[height=\outsize]{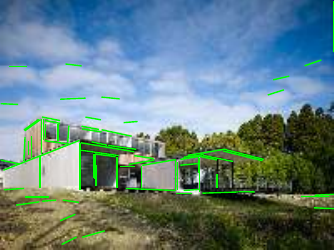}
\includegraphics[height=\outsize]{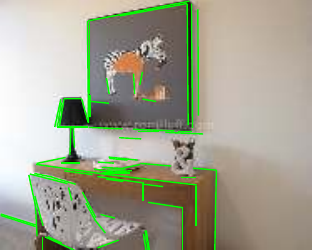}
\includegraphics[height=\outsize]{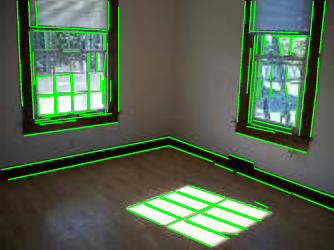}

\hspace{\xshift}
\begin{tikzpicture}
\node[rotate=90] (Ours) at (0,0) {DWP};
\draw[opacity=0] (-\outsize*0.01,-\outsize*0.5) rectangle (\outsize*0.01,\outsize*0.5);
\end{tikzpicture}
\includegraphics[height=\outsize]{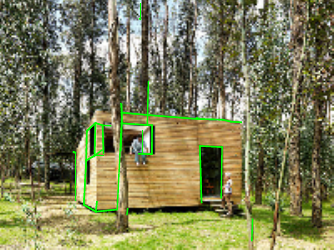}
\includegraphics[height=\outsize]{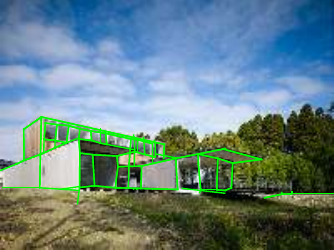}
\includegraphics[height=\outsize]{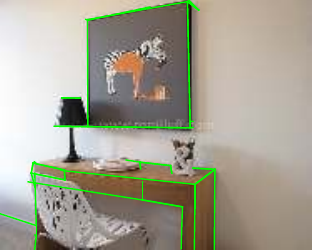}
\includegraphics[height=\outsize]{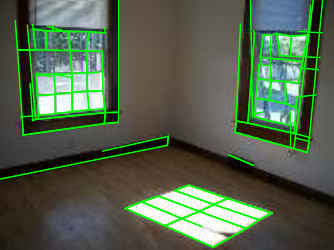}

\hspace{\xshift}
\begin{tikzpicture}
\node[rotate=90] (Ours) at (0,0) {AFM};
\draw[opacity=0] (-\outsize*0.01,-\outsize*0.5) rectangle (\outsize*0.01,\outsize*0.5);
\end{tikzpicture}
\includegraphics[height=\outsize]{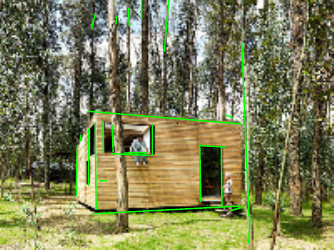}
\includegraphics[height=\outsize]{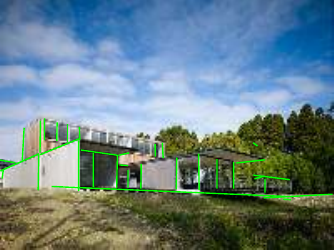}
\includegraphics[height=\outsize]{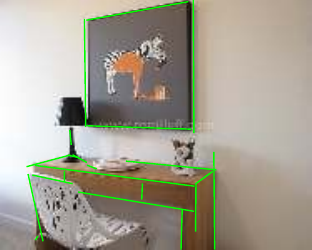}
\includegraphics[height=\outsize]{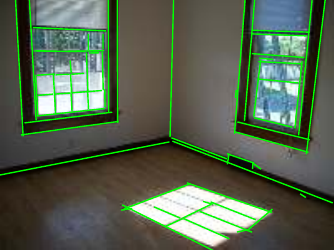}

\hspace{\xshift}
\begin{tikzpicture}
\node[rotate=90] (Ours) at (0,0) {AFM++};
\draw[opacity=0] (-\outsize*0.01,-\outsize*0.5) rectangle (\outsize*0.01,\outsize*0.5);
\end{tikzpicture}
\includegraphics[height=\outsize]{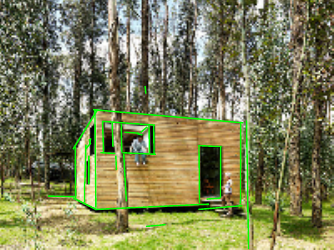}
\includegraphics[height=\outsize]{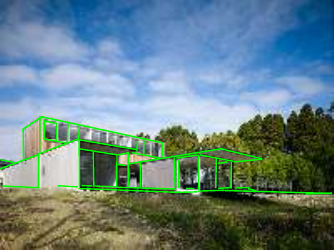}
\includegraphics[height=\outsize]{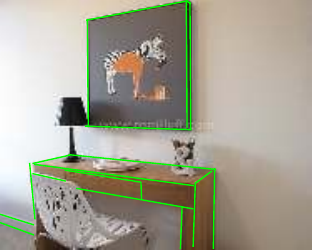}
\includegraphics[height=\outsize]{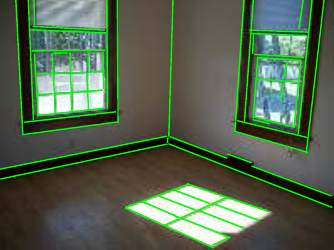}

\hspace{\xshift}
\begin{tikzpicture}
\node[rotate=90] (GT) at (0,0) {GT};
\draw[opacity=0] (-\outsize*0.01,-\outsize*0.5) rectangle (\outsize*0.01,\outsize*0.5);
\end{tikzpicture}
\includegraphics[height=\outsize]{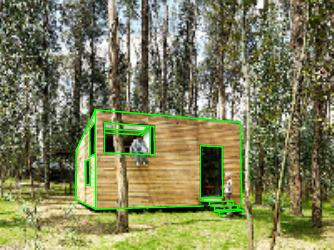}
\includegraphics[height=\outsize]{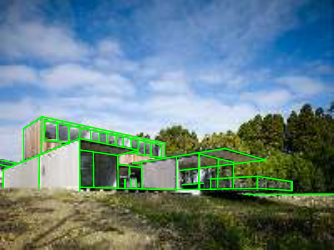}
\includegraphics[height=\outsize]{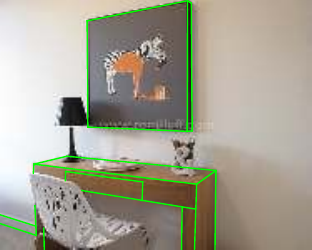}
\includegraphics[height=\outsize]{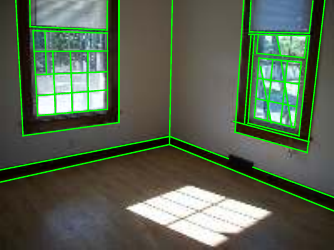}
\caption{Some results of line segment detection of different approaches on the Wireframe \cite{Huang2018a} dataset. From top to bottom: LSD \cite{VonGioi2010}, MCMLSD \cite{Almazan_2017_CVPR}, Linelet \cite{Cho2018}, DWP~\cite{Huang2018a}, AFM~\cite{AFMCVPR} with the \emph{a-trous} Residual U-Net and AFM++ proposed in this paper. 
}
\label{fig:results_wireframe}
\vspace{-3mm}
\end{figure*}
	\begin{figure*}[ht!]
\centering
\def\outsize{0.13\textheight}
\def\xshift{-4mm}
\hspace{\xshift}
\begin{tikzpicture}
\node[rotate=90] (Ours) at (0,0) {LSD};
\draw[opacity=0] (-\outsize*0.01,-\outsize*0.5) rectangle (\outsize*0.01,\outsize*0.5);
\end{tikzpicture}
\includegraphics[height=\outsize]{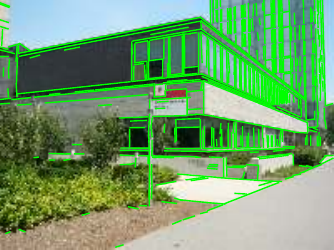}
\includegraphics[height=\outsize]{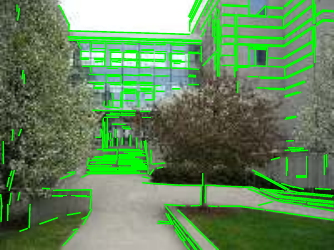}
\includegraphics[height=\outsize]{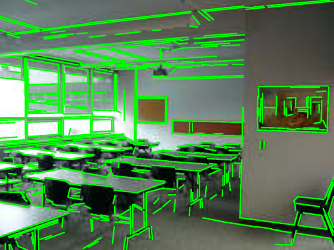}
\includegraphics[height=\outsize]{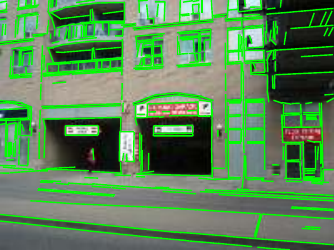}
\\
\hspace{\xshift}
\begin{tikzpicture}
\node[rotate=90] (Ours) at (0,0) {MCMLSD};
\draw[opacity=0] (-\outsize*0.01,-\outsize*0.5) rectangle (\outsize*0.01,\outsize*0.5);
\end{tikzpicture}
\includegraphics[height=\outsize]{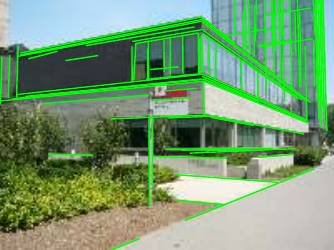}
\includegraphics[height=\outsize]{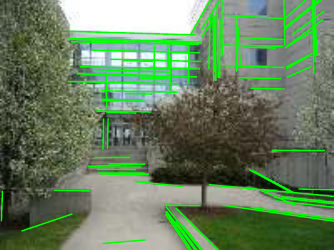}
\includegraphics[height=\outsize]{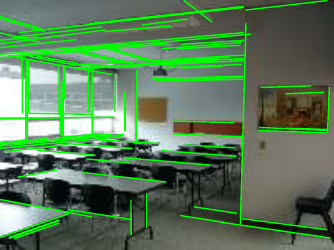}
\includegraphics[height=\outsize]{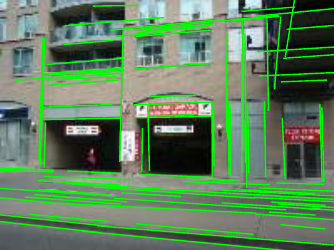}
\\
\hspace{\xshift}
\begin{tikzpicture}
\node[rotate=90] (Ours) at (0,0) {Linelet};
\draw[opacity=0] (-\outsize*0.01,-\outsize*0.5) rectangle (\outsize*0.01,\outsize*0.5);
\end{tikzpicture}
\includegraphics[height=\outsize]{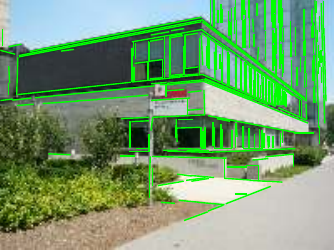}
\includegraphics[height=\outsize]{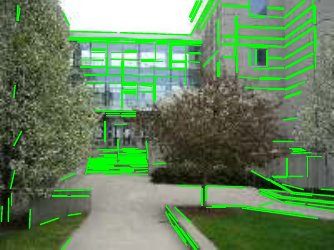}
\includegraphics[height=\outsize]{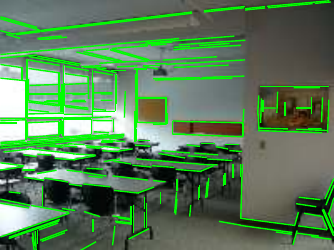}
\includegraphics[height=\outsize]{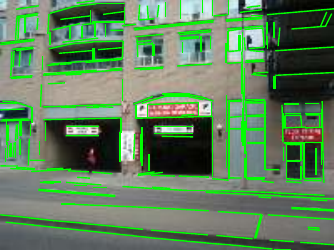}

\hspace{\xshift}
\begin{tikzpicture}
\node[rotate=90] (Ours) at (0,0) {DWP};
\draw[opacity=0] (-\outsize*0.01,-\outsize*0.5) rectangle (\outsize*0.01,\outsize*0.5);
\end{tikzpicture}
\includegraphics[height=\outsize]{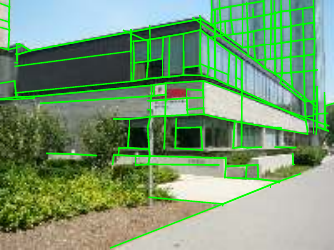}
\includegraphics[height=\outsize]{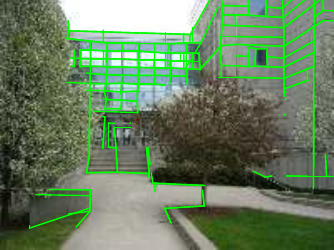}
\includegraphics[height=\outsize]{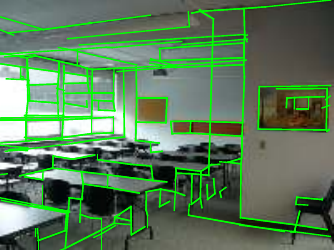}
\includegraphics[height=\outsize]{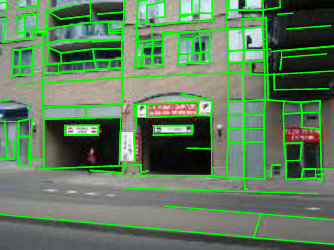}

\hspace{\xshift}
\begin{tikzpicture}
\node[rotate=90] (Ours) at (0,0) {AFM};
\draw[opacity=0] (-\outsize*0.01,-\outsize*0.5) rectangle (\outsize*0.01,\outsize*0.5);
\end{tikzpicture}
\includegraphics[height=\outsize]{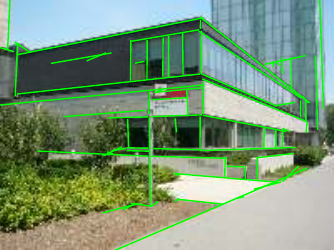}
\includegraphics[height=\outsize]{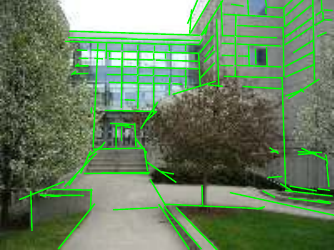}
\includegraphics[height=\outsize]{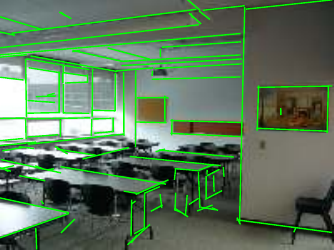}
\includegraphics[height=\outsize]{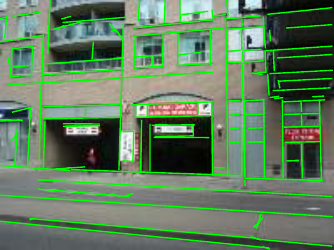}

\hspace{\xshift}
\begin{tikzpicture}
\node[rotate=90] (Ours) at (0,0) {AFM++};
\draw[opacity=0] (-\outsize*0.01,-\outsize*0.5) rectangle (\outsize*0.01,\outsize*0.5);
\end{tikzpicture}
\includegraphics[height=\outsize]{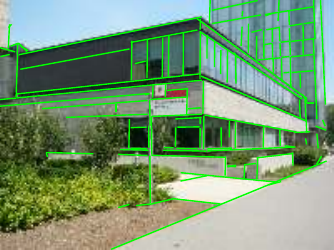}
\includegraphics[height=\outsize]{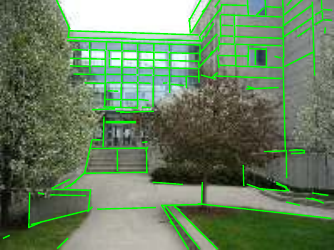}
\includegraphics[height=\outsize]{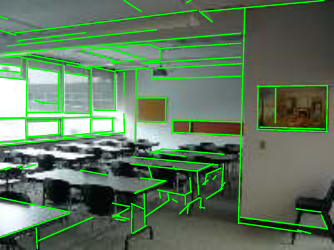}
\includegraphics[height=\outsize]{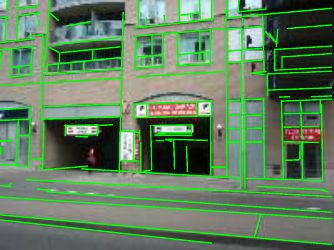}

\hspace{\xshift}
\begin{tikzpicture}
\node[rotate=90] (GT) at (0,0) {GT};
\draw[opacity=0] (-\outsize*0.01,-\outsize*0.5) rectangle (\outsize*0.01,\outsize*0.5);
\end{tikzpicture}
\includegraphics[height=\outsize]{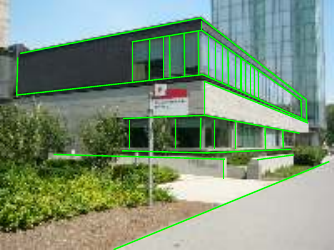}
\includegraphics[height=\outsize]{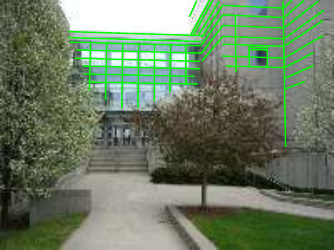}
\includegraphics[height=\outsize]{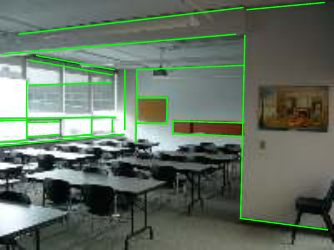}
\includegraphics[height=\outsize]{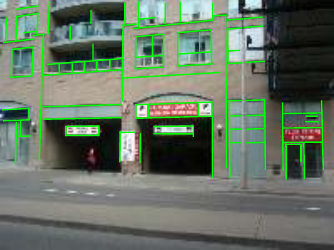}

\caption{Some results of line segment detection of different approaches on the YorkUrban \cite{Denis2008} datasets. From top to bottom: LSD \cite{VonGioi2010}, MCMLSD \cite{Almazan_2017_CVPR}, Linelet \cite{Cho2018}, DWP \cite{Huang2018a}, AFM~\cite{AFMCVPR} with the \emph{a-trous} Residual U-Net and AFM++ proposed in this paper. 
}
\label{fig:results_york}
\vspace{-3mm}
\end{figure*}
	\paragraph*{Precision \& Recall}
	We first evaluate the proposed method on the Wireframe dataset~\cite{Huang2018a}. The precision-recall curves and the F-measure are presented in Figure~\ref{fig:pr-curve-wireframe} and Table~\ref{tab:fmeasure-and-fps}, respectively. As is shown, the proposed AFM++ sets a new state-of-the-art performance, that is the F-measure of $0.823$. This achievement is dramatically better than DWP~\cite{Huang2018a}, with a performance improvement of approximately $10$ percent. Compared with the previous version AFM~\cite{AFMCVPR}, AFM++ improves the F-measure by $5$ percent on this dataset. This demonstrates the usefulness of outlier removal module and better optimizer, which will be further discussed below.
	
	Furthermore, we also evaluate our proposed approach on the YorkUrban dataset~\cite{Denis2008} and the performance comparison is given in Table~\ref{tab:fmeasure-and-fps} and Figure~\ref{fig:pr-curve-york}. Consistent with the results on the Wireframe dataset, our work (AFM and AFM++) beats those representative algorithms by a large margin. In particular, AFM++ achieves an F-measure of $0.672$, advancing the state-of-the-art performance by $4.5$ percent (over $0.627$ reported by DWP~\cite{Huang2018a}). Note that the YorkUrban dataset only focuses on the Manhattan frame estimation, which results in that some line segments in the images are not labeled. Therefore, one may observe that the performance on the YorkUrban dataset is generally lower than that on the Wireframe dataset.
	
	\paragraph*{Visualization and Discussion}
	We visualize the line segments detected by different methods in Figure~\ref{fig:results_wireframe} for the Wireframe dataset and Figure~\ref{fig:results_york} for the YorkUrban dataset, respectively.
	The threshold configurations for visualization are as follows:
	\begin{enumerate}
		\item The \emph{a-contrario} validation of LSD and Linelet are set to $-\log\epsilon = 0.01\cdot1.75^{8}$;
		\item The top $90$ line segments detected by MCMLSD are visualized;
		\item The threshold of the line heat map is $10$ for DWP;
		\item The aspect ratio is set to $0.2$ for AFM~\cite{AFMCVPR} and AFM++.
	\end{enumerate}
	
	As we can see from Figure~\ref{fig:results_wireframe} and Figure~\ref{fig:results_york}, the deep learning based approaches, including AFM++, AFM~\cite{AFMCVPR} and DWP~\cite{Huang2018a}, generally perform better on the two datasets than the other approaches, including LSD~\cite{VonGioi2010}, MCMLSD~\cite{Almazan_2017_CVPR} and Linelet~\cite{Cho2018}, since they utilize the global information to capture the low-contrast regions while suppressing the false detections in the edge-like texture regions. The approaches~\cite{VonGioi2010,Almazan_2017_CVPR,Cho2018} 
	only infer line segments from local features, thus causing incomplete detection results and a number of false detections even with powerful validation processes.
	Although the overall F-measure of LSD~\cite{VonGioi2010} is slightly better than Linelet~\cite{Cho2018}, the qualitative visualizations of Linelet~\cite{Cho2018} are cleaner.
	
	For deep learning based approaches, AFM++ significantly outperforms AFM~\cite{AFMCVPR} and DWP~\cite{Huang2018a} with fewer false detections and more accurate line segment localization. Compared with AFM, we are able to resolve the overshooting issue in the endpoint estimation because of the better regional attraction learning and outlier removal module. In contrast to DWP~\cite{Huang2018a}, AFM and AFM++ get rid of junction detection and line heat map prediction, thus resolving the local ambiguity for line segment detection in an efficient way. Since DWP~\cite{Huang2018a} requires junctions for line segment detection, the results are not well localized by the inaccurately estimated orientation of junctions. Besides, the incorrect edge pixels will mislead the merging module in DWP~\cite{Huang2018a} to generate false detections by mistakenly connecting some junction pairs.
	
	\paragraph*{Inference Speed}
	We compare the inference speed of the aforementioned algorithms on the Wireframe dataset. The time cost is calculated over the entire testing dataset and the average frames-per-second (FPS) is reported in the last column of Table~\ref{tab:fmeasure-and-fps}.
	All the experiments were conducted on a PC workstation  equipped with an Intel Xeon E5-2620 $2.10$ GHz CPU and $4$ NVIDIA Titan X GPU devices. Only one GPU is used and the CPU programs are executed in a single thread. 
	
	As reported in Table \ref{tab:fmeasure-and-fps}, in addition to the state-of-the-art performance, our method is also computationally inexpensive. 
	Benefiting from the simplicity of our novel formulation, AFM-based methods run faster than all the other methods except for LSD. AFM (U-Net) is the fastest among the AFM based approaches, and is the second compared to LSD.
	DWP~\cite{Huang2018a} spends much time for junction and line heat map merging. Meanwhile, our method resizes the input images into $320\times 320$ and then transforms the output line segments to the original size without loss of information, which further reduces the computational cost. Compared with AFM~\cite{AFMCVPR}, the outlier removal module in AFM++ only retains well-estimated attraction vectors, which also improves the computational speed.
	
	\begin{figure}[h]
		\centering
		\subfigure[Results by the model trained on $320\times 320$ samples]{
			\includegraphics[height=0.12\textheight]{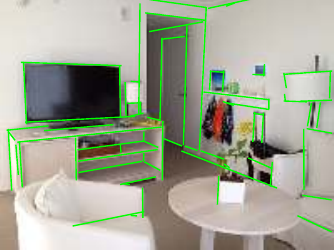}
			\includegraphics[height=0.12\textheight]{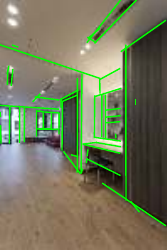}
			\includegraphics[height=0.12\textheight]{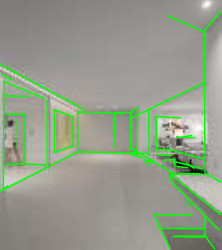}
		}	
		\vspace{-3mm}
		\subfigure[Results by the model trained on $512\times 512$ samples]{
			\includegraphics[height=0.12\textheight]{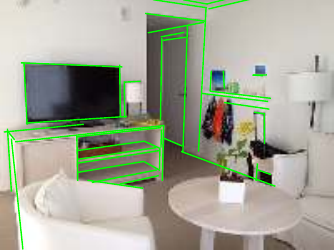}
			\includegraphics[height=0.12\textheight]{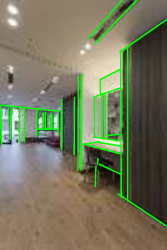}
			\includegraphics[height=0.12\textheight]{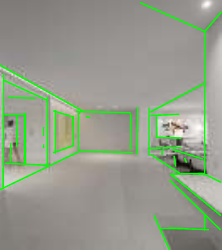}
		}	
		\vspace{-3mm}
		\caption{Line segments detected on images of different resolutions.}
		\label{fig:diff-resolution}
		\vspace{-3mm}
	\end{figure}
	\begin{figure*}
		\centering
		\raisebox{0.11\linewidth}{\rotatebox[origin=c]{90}{Images}}
		\includegraphics[width=0.23\linewidth]{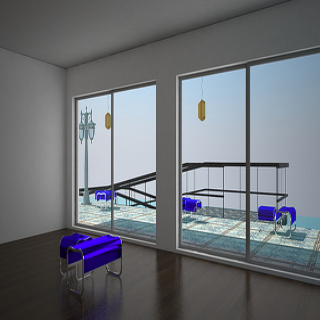}
		\includegraphics[width=0.23\linewidth]{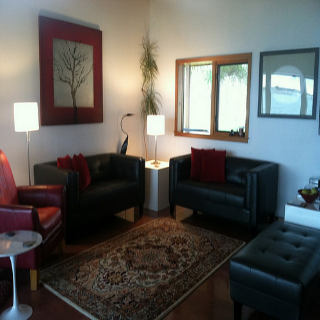}
		\includegraphics[width=0.23\linewidth]{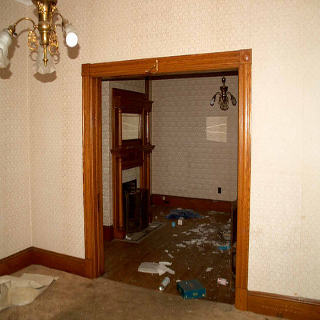}
		\includegraphics[width=0.23\linewidth]{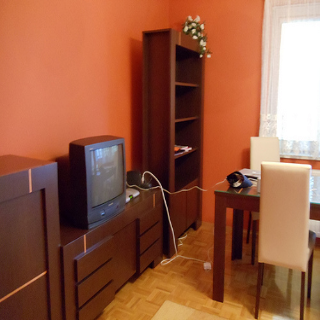}
		
		\vspace{1mm}
		
		\raisebox{0.11\linewidth}{\rotatebox[origin=c]{90}{Saliency Maps}}
		\includegraphics[width=0.23\linewidth]{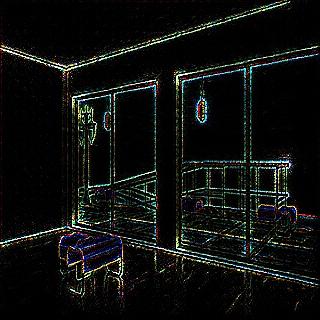}
		\includegraphics[width=0.23\linewidth]{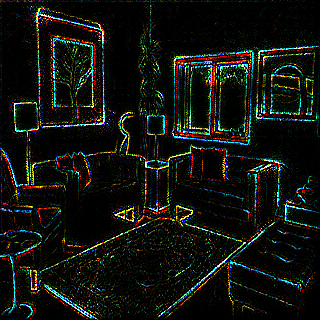}
		\includegraphics[width=0.23\linewidth]{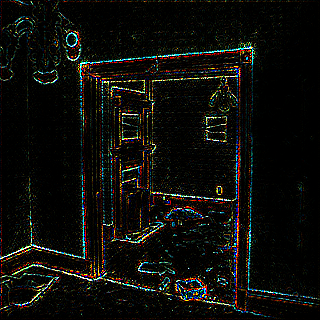}
		\includegraphics[width=0.23\linewidth]{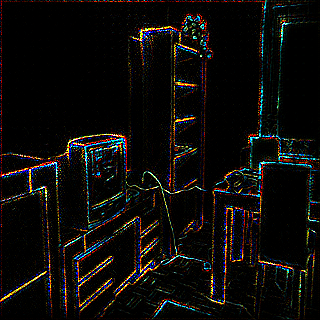}
		
		\caption{Visualized interpretation of the learned network. The top row displays some examples of images and the bottom row displays corresponding saliency maps obtained by Guided Backpropagation~\cite{Guided-BP}.}
		\label{fig:interpretation}
	\end{figure*}
	\subsection{Interpretability}
	In this section, we are going to discuss what are the network learned. Generally speaking, the learning target of our network is the attraction field map, however, it is hard to understand the learning process by simply observing the predictions of attraction field map. 
	Alternatively, we use Guided Backpropagation~\cite{Guided-BP} to visualize which pixels are important for the attraction field prediction and line segment detection. 
	Guided Backpropagation~\cite{Guided-BP} interprets pixels' importance for prediction by calculating the gradient flow from the prediction layer to the input images. The magnitude of the gradients flowed back to the input image indicates the change of the pixels that will affect the final prediction. Different from the vanilla backpropogation, Guided Backpropagation only retains the positive gradients in the ReLU layer and passes the modified gradients to the previous layer. The gradients flowed to the input images are used for visualization. As discussed in~\cite{Guided-BP}, the gradients with positive values indicate corresponding pixels with high influence for prediction. Accordingly, we use the positive gradient maps (with respect to the input image) as the saliency maps for visualization. In the computation, the gradients for the last layer are set to $1$. As shown in Figure~\ref{fig:interpretation}, it is interesting to see that the learned network will automatically perceive the geometric structures of the input image.
	This visualization results could help us to understand why the convolutional neural networks can be used for line segment detection.

	\subsection{Tweaks and Discussion for Further Improvements}
	\begin{table}[t!]
		\centering
		\caption{Performance change by increasing image resolution, using a better optimization method and adding the outlier removal module.}
		\vspace{-3mm}
		\footnotesize
		\label{tab:fmeasure-resolution}
		\resizebox{\linewidth}{!}{
			\begin{tabular}{|c|c|c|c|c|c|}
				\hline
				\makecell{Backbone/\\Optimizer}& resolution & \makecell{Outlier\\removal} & \makecell{Wireframe\\ dataset} & \makecell{YorkUrban\\ dataset} & { FPS}  \\
				\hline
				\hline
				\multirow{4}{*}{
					\makecell{\emph{a-trous}/\\SGD}
				} & $320\times 320$ & w/o & 0.774 & 0.647 &  6.6 \\
				& $512\times 512$ & w/o & 0.794 & 0.660 &  2.3\\
				& $320\times 320$ & w   & 0.807 & 0.663 &  8.0\\
				& $512\times 512$ & w   & 0.826 & 0.674 &  3.5\\
				\hline
				\hline
				\multirow{4}{*}{
					\makecell{\emph{a-trous}/\\Adam}
				} 
				& $320\times 320$ & w/o & 0.792 & 0.659 &  6.6 \\
				& $512\times 512$ & w/o & 0.802 & 0.665 &  2.3\\
				& $320\times 320$ & w   & 0.823 & 0.672 &  8.0\\
				& $512\times 512$ & w   & \textbf{0.831} &  \textbf{0.680} &  3.5\\
				\hline
			\end{tabular}
		}
	\end{table}

	In this section, we explore how to further improve the performance with some useful tweaks. In detail, we train the \emph{a-trous} Residual U-Net with higher resolution images ($512\times 512$) and different optimization methods. Furthermore, the outlier removal module with statistical priors  is verified to be effective for line segment detection. By combining these tweaks, we obtain a higher F-measure, that is, $0.831$ on the Wireframe dataset. 
	
	\paragraph*{Training with Higher Resolutions}
	Since we adopt the encoder-decoder architecture for the attraction field learning, it is interesting to determine whether higher-resolution samples are conducive to extracting finer features for AFM learning. 
	In this experiment, we increase the resolution of the training samples from $320\times320$ (default setting) to $512\times 512$ for training and testing, while keeping the other settings the same as the previous configuration.
	The results are reported in Table~\ref{tab:fmeasure-resolution}.
	
	Generally speaking, increasing the resolution of training samples increases the accuracy of LSD evidently. For example, when using \emph{a-trous} as the backbone and stochastic gradient descent (SGD) as the optimizer, the F-measure is improved by about $2$ percent (from $0.773$ to $0.794$).
	For qualitative evaluations, some results of the line segments detected with different resolution samples ($320\times 320$, $512\times 512$) are plotted in Figure~\ref{fig:diff-resolution}.  As is shown, the higher resolution of the training samples is, more complete results with fewer false detection rate are obtained. However, the increased image size will slow down the inference speed.
	
	\paragraph*{Better Optimization Method}
	It has been demonstrated that the Adam  optimizer performs better than SGD in image classification~\cite{Adam-Optimizer}.
	Inspired by this, we use the Adam optimizer to optimize the model rather than SGD adopted in AFM~\cite{AFMCVPR}. As shown in Table~\ref{tab:fmeasure-resolution}, the Adam optimizer improves the F-measure by $2$ percent on the Wireframe dataset compared with SGD. The Adam optimizer will not increase the computational cost in testing phase.
	
	\paragraph*{Outlier Removal}
	Although the proposed regional attraction uses all the learned attraction vectors, the ConvNets cannot ensure that the vectors in each pixel can be predicted as accurately as possible. Besides, our numerical stable normalization in Equation~\eqref{eq:log-normalize} will implicitly give the attraction vectors with smaller magnitude a large penalty. Therefore, the outlier removal module with statistical priors can filter out the inaccurately estimated attraction vectors in an efficient way. In this experiment, we filter out the attraction vectors of which the $\ell_2$ norm is greater than $\gamma = 0.02\times \min(H,W)$. Recall that $H$ and $W$ are the size of training samples. As reported in Table~\ref{tab:fmeasure-resolution}, the outlier removal improves the F-measure performance using the same training resolution and optimizer. 
	Meanwhile, the outlier removal can reduce the amount of attraction vectors for the squeeze module and slightly improves the inference speed.
	
	\section{Conclusion and Further Work}\label{sec:conclusion}
	In this paper, we proposed a method of representing and characterizing the 1D geometry of line segments by using all pixels in the image lattice. The problem of line segment detection (LSD) is then posed as a problem of region coloring which is addressed by learning convolutional neural networks. The region coloring formulation of LSD harnesses the best practices developed in deep learning based semantic segmentation methods such as the encoder-decoder architecture and the {\em a-trous} convolution.
	In the experiment, our method is tested on two widely used LSD benchmarks, \ie, the Wireframe~\cite{Huang2018a} and the YorkUrban~\cite{Denis2008} datasets, with state-of-the-art performance obtained in both accuracy and speed. 
	
	In the future, we will exploit how to simultaneously detect line segments and junctions together in a convolutional neural network. Considering the simplicity and superior performance, we hope that the new perspective provided in this work can facilitate and motivate better line segment detection and geometric scene understanding. 
	Furthermore, we will study the application of the proposed line segment detector to many up-level vision tasks such as Structure-from-Motion (SfM), SLAM and single-view 3D reconstruction.
	
	\section*{Acknowledgements}
	This work was supported by the National Natural Science Foundation of China under Grant 61922065, Grant 61771350 and Grant 41820104006. This work was also supported in part by EPSRC grant Seebibyte EP/M013774/1 and EPSRC/MURI grant EP/N019474/1. Nan Xue was also supported by China Scholarship Council. T. Wu was supported in part by ARO Grant W911NF1810295 and NSF IIS-1909644. The views presented in this paper are those of the authors and should not be interpreted as representing any funding agencies.
	\bibliographystyle{IEEEtran}
	\bibliography{ref}
	
\end{document}